%% file: main.tex
\documentclass[10pt,twocolumn,letterpaper]{article}

\usepackage[pagenumbers]{cvpr} 
\usepackage{pifont}  \newcommand{\cmark}{\textcolor{green!50!black}{\ding{51}}}  \newcommand{\xmark}{\textcolor{red!70!black}{\ding{55}}}
\input{preamble}

\definecolor{cvprblue}{rgb}{0.21,0.49,0.74}
\usepackage[pagebackref,breaklinks,colorlinks,allcolors=cvprblue]{hyperref}

\def\confName{CVPR}
\def\confYear{2027}

\title{TERZI: \underline{T}hree-dimensional Garment G\underline{e}neration and \underline{R}etargeting via Optimi\underline{z}ation-free \underline{I}nference}

\author{Emirhan Bilgi\c{c} \qquad R{\i}za Alp G\"{u}ler\\
Department of Computing, Imperial College London
}

\begin{document}
\twocolumn[{%
\renewcommand\twocolumn[1][]{#1}%
\maketitle
\begin{center}
  \includegraphics[width=\linewidth]{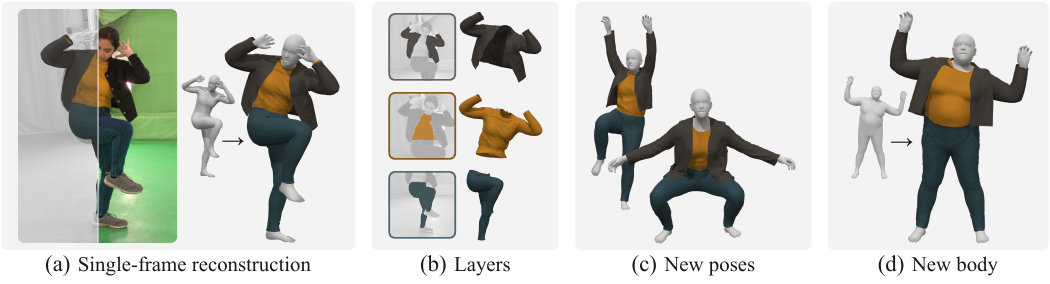}
  \captionof{figure}{\textbf{\method uses a single image of a clothed person and an estimate of their body to reconstruct image-aligned 3D garment meshes (a).}
  Selecting each garment with its image mask yields separate meshes for the outer, upper, and lower layers \textbf{(b)}.
  Keeping the image and masks fixed, we can generate the same garments directly on the body in new poses \textbf{(c)}.
  Our system can also generate the same garments on a target body with a different shape and pose, without explicit training for cross-body transfer \textbf{(d)}.}
  \label{fig:teaser}
\end{center}
\vspace{1em}
}]
\input{sec/0_abstract}    
\input{sec/1_intro}
\input{sec/2_related}
\input{sec/3_method}
\input{sec/4_experiments}
\input{sec/5_conclusion}
{
    \small
    \bibliographystyle{ieeenat_fullname}
    \bibliography{main}
}

\input{sec/X_suppl}

\end{document}

%% file: preamble.tex
\newcommand{\method}{TERZI\xspace}
\usepackage{multirow}
\usepackage{makecell}


%% file: sec/0_abstract.tex
\begin{abstract}
Virtual try-on, animation and garment retargeting need clothing as an independent asset, yet single-image methods either fuse body and clothing into one surface or recover garments through per-subject optimization taking 10--46 minutes per image followed by post-hoc collision fixes.
We present TERZI, a generative framework that reconstructs decomposable 3D garment meshes conditioned on a given parametric body model and on garment appearance extracted from a single image.
Two body-conditioned flow-matching transformers, operating in a sparse structured 3D latent space, first predict the cells the garment occupies, then the surface geometry inside them.
The garments are separate from the conditioning body by construction, and image masks select an individual garment or the complete outfit.
The image determines which garment is generated, and the body determines its pose and shape.
With the body estimated from the image, the model reconstructs the garment; with a different body, the same model generates the garment directly in the new pose or body shape, without skinning or simulation.
A differentiable penetration penalty between the decoded garment surface and the conditioning body makes the generator itself collision-aware, so no separate collision-correction stage is needed at inference.
Trained only on synthetic data, our method generalizes to real, in-the-wild images and achieves state-of-the-art accuracy on generation and retargeting (garment Chamfer distance 1.48\,cm \vs 1.88\,cm for the strongest baseline on 4D-DRESS).
A distilled one-step student generates an outfit in 0.1\,s, with comparable accuracy. We commit to releasing the training code, model weights, and config files for full reproducibility.
Project page: \url{https://emirhanbilgic.github.io/TERZI/}
\end{abstract}

%% file: sec/1_intro.tex
\vspace{-1.0em}
\section{Introduction}
\label{sec:intro}

3D cloth and human body reconstruction is an essential task for applications like virtual try-on, retargeting to a new body, simulation, animation and avatar editing. And all of them need the clothing of a person as separate 3D garments, not only as part of a single surface.
Given a single RGB image of a clothed person, we therefore aim to recover a \emph{decomposed} 3D representation: a body mesh and one or more garment meshes as separate surfaces that are image-aligned and do not interpenetrate.
Most clothed-human reconstruction methods output the body and its clothing as a single surface~\cite{saito2019pifu,saito2020pifuhd,xiu2022icon,xiu2023econ}, which is sufficient for viewing but not for these applications.
The few methods that do decompose~\cite{li2023isp,nam2025declot,vuran2025remu} either rely on per-subject optimization (10--46\,min per image) or on template and sewing-pattern priors that restrict the garments they can express, typically resolve body--garment collisions in a post-hoc step, and/or are not image-aligned.
Native 3D generators with sparse structured latents~\cite{xiang2025trellis,xiang2025trellis2} produce detailed, arbitrary-topology geometry, but they model objects in isolation, with no notion of a body to dress, garment to animate, or an image to match.

Our method, \method, formulates garment reconstruction as \emph{body-conditioned generation}: the garment is synthesized around a given body, guided by the image and selected by a mask (\cref{fig:pipeline}).
Given an image crop, an SMPL-X body~\cite{pavlakos2019smplx} estimated by an off-the-shelf human mesh recovery (HMR) system and an optional garment mask, two flow-matching transformers (DiTs), initialized from TRELLIS.2~\cite{xiang2025trellis2}, generate the garment, first the sparse cells it occupies (SS DiT), then its surface inside them (SLat DiT).
We generate only the garment, not the body: a jointly generated body can be implausible, and the garment inherits its errors, whereas mature HMR methods~\cite{shen2024gvhmr,wang2025prompthmr,sam3dbody2025} already provide accurate bodies.
Keeping the image fixed and changing the body's pose or shape generates the same garments in a new pose or on a different body, with the same trained model.

Compared with single-surface methods, our method returns the body and each garment as separate meshes; compared to optimization-based decomposition, it does not require a test-time adjustment; and compared to template and sewing-pattern methods, it is not restricted to predefined garment categories.
Trained on synthetic BEDLAM2.0~\cite{tesch2025bedlam2} data, it generalizes to different datasets (4D-DRESS~\cite{wang20244ddress} in \cref{tab:main}, and CloSe~\cite{antic2024close} in the supplementary material), and real, in-the-wild images (RICH~\cite{huang2022rich} and 3DPW~\cite{vonmarcard2018recovering}, \cref{tab:wild}) with hard poses and is more accurate than both single-surface and decomposed methods (\cref{tab:main}).
Our contributions are:
\begin{itemize}[leftmargin=*,itemsep=1pt,topsep=2pt]
  \item \textbf{Feed-forward, body-conditioned garment generation.} A two-stage flow-matching model generates individual garments or complete outfits from an image, a body and an optional mask in a single pass; a distilled student takes about 0.1\,s per image, compared with 10--46\,min for optimization-based methods.
  \item \textbf{Generation and retargeting in a single model.} Conditioning on images from other frames of the same sequence during training teaches the model to read the garment from the image and its pose and shape from the body, so it transfers garments to new bodies and poses without skinning or simulation and is more accurate than animating the ground-truth garment (\cref{tab:animation}).
  \item \textbf{Collision-aware fine-tuning.} A differentiable penetration penalty on the decoded garment surface, anchored to the frozen model, teaches the generator to keep garments outside the body, with no collision correction at inference (\cref{sec:penetration}).
\end{itemize}

%% file: sec/2_related.tex
\section{Related Work}
\label{sec:related}

\begin{table}[t]
  \centering
  \caption{\textbf{Capabilities of clothed-human and garment reconstruction methods.}
  \emph{Template-free}: not limited to a fixed set of garment templates or sewing-pattern categories.
  \emph{Retargeting}: the method itself reposes garment on a new pose (re-posing by plain linear blend skinning or an external cloth simulator is not counted).}
  \label{tab:capabilities}
  \setlength{\tabcolsep}{3pt}
  \resizebox{\linewidth}{!}{%
  \begin{tabular}{@{}lcccccccc@{}}
    \toprule
    Method & \makecell{Feed-\\forward} & \makecell{Template-\\free} & \makecell{Image-\\aligned} & \makecell{Separate\\garment} & \makecell{Layered\\garments} & \makecell{Penetration-\\aware} & \makecell{Re-\\targeting} \\
    \midrule
    ReMu~\cite{vuran2025remu}              & \xmark & \cmark & \cmark & \cmark & \cmark & \cmark & \xmark \\
    TRELLIS.2~\cite{xiang2025trellis2}     & \cmark & \cmark & \xmark & \xmark & \xmark & \xmark & \xmark \\
    HumanNOVA~\cite{hu2026humannova}       & \cmark & \cmark & \xmark & \xmark & \xmark & \xmark & \xmark \\
    SiTH~\cite{ho2024sith}                 & \cmark & \cmark & \cmark & \xmark & \xmark & \xmark & \xmark \\
    ISP~\cite{li2023isp}                   & \xmark & \xmark & \cmark & \cmark & \cmark & \cmark & \cmark \\
    ChatGarment~\cite{bian2025chatgarment} & \cmark & \xmark & \xmark & \cmark & \xmark & \xmark & \xmark \\
    DeClotH~\cite{nam2025declot}      & \xmark & \xmark & \cmark & \cmark & \xmark & \cmark & \xmark \\
    Dress-1-to-3~\cite{li2025dress1to3} & \xmark & \xmark & \cmark & \cmark & \xmark & \cmark & \cmark \\
    \midrule
    \textbf{\method (ours)}                & \cmark & \cmark & \cmark & \cmark & \cmark & \cmark & \cmark \\
    \bottomrule
  \end{tabular}}
\end{table}

\begin{figure*}[t]
  \centering
  \includegraphics[width=\linewidth]{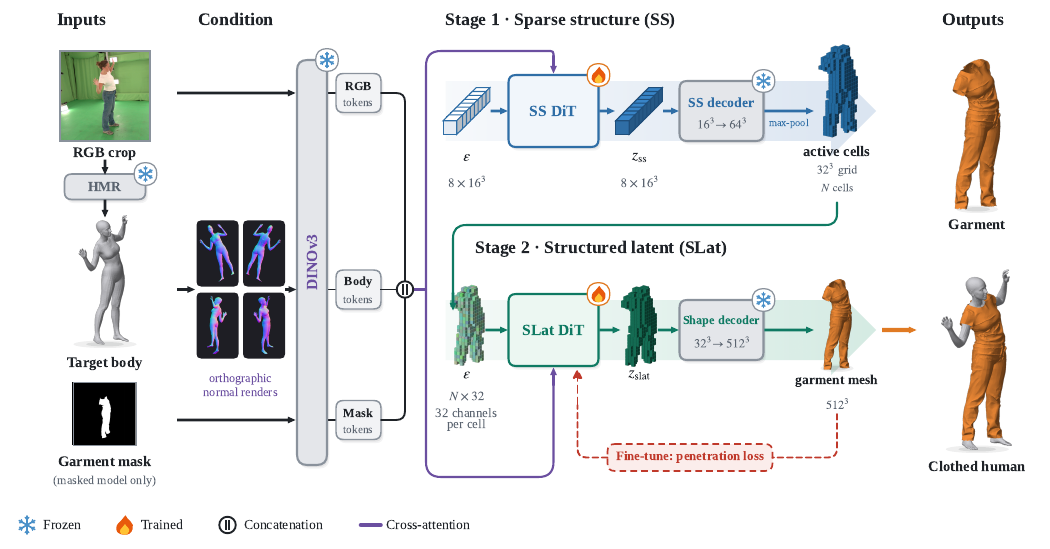}
  \caption{\textbf{Overview of \method.}
    The image $I$, the body $B$ estimated by an HMR method and, a garment mask $M$ (only for the masked model) are encoded by a frozen DINOv3 into tokens that serve as cross-attention context in every block of two flow-matching DiTs, which generate the garment $G$ around $B$.
    \emph{Stage~1} generates the structure latent $z_\text{ss}$, which the frozen structure decoder turns into the active cells the garment occupies.
    \emph{Stage~2} generates the structured latent $z_\text{slat}$ on these cells, which the frozen shape decoder turns into the garment mesh.
    Body and garment are separate meshes by construction.
    \emph{Training vs.\ inference:} during training, each ground-truth garment is placed in the image-aligned cube and encoded by the frozen encoders into $z_\text{ss}$ and $z_\text{slat}$, the denoising targets of the two stages. Training data is mixed to enable retargeting (see \cref{fig:train_mix}). A later fine-tuning stage trains the Stage~2 DiT with a penetration loss on the decoded garment.  At inference, both latents are generated from noise and only the decoders are used. No penetration fix is needed during inference. 
}
  \label{fig:pipeline}
\end{figure*}

\Cref{tab:capabilities} summarizes how our method differs from representative prior methods.

\subsection{Single-Surface Reconstruction and Generation}
These methods reconstruct the person and their clothing as one surface. ICON~\cite{xiu2022icon}, ECON~\cite{xiu2023econ} and SIFU~\cite{zhang2024sifu} condition them on an SMPL-X body~\cite{loper2015smpl,pavlakos2019smplx}, and SiTH~\cite{ho2024sith} and PSHuman~\cite{li2025pshuman} first hallucinate unseen views with diffusion, taking up to two minutes per image.
HumanNOVA~\cite{hu2026humannova} decodes a triplane avatar around a coarse SMPL prior feed-forward, and native 3D generators such as TRELLIS~\cite{xiang2025trellis} and TRELLIS.2~\cite{xiang2025trellis2} produce arbitrary-topology assets from a structured sparse latent with flow-matching transformers~\cite{lipman2023flow,liu2023rectified,peebles2023dit}.
Our method builds on the latent space of TRELLIS.2 but conditions the generator on the body and the image, so each garment is its own image-aligned mesh around the given body, with collision avoidance learned rather than fixed afterwards, and it is more accurate than the segmented single-surface outputs (\cref{tab:main}).

\subsection{Decomposed Reconstruction}
Optimization-based methods fit separate garments to each image.
ReMu~\cite{vuran2025remu} reconstructs one clothed human per garment layer; DeClotH~\cite{nam2025declot} optimizes cloth and body meshes with template regularization; and Dress-1-to-3~\cite{li2025dress1to3} refines a predicted sewing pattern, in up to two hours per outfit.
ISP~\cite{li2023isp} maps 2D sewing panels onto the body and optimizes each garment; its successor adds normal-fitted deformation priors~\cite{li2024garmentrecovery}.
Earlier template methods fit category-specific garment models~\cite{jiang2020bcnet,corona2021smplicit,moon2022clothwild}.
Templates and patterns cover only the categories they were built for, and per-sample fitting takes minutes (10.4\,min for ReMu and 46\,min for ISP).
Pattern generators predict sewing patterns from images~\cite{liu2023sewformer} or text~\cite{he2024dresscode}, or with multimodal models over programs~\cite{korosteleva2023garmentcode}; ChatGarment~\cite{bian2025chatgarment} fine-tunes a vision--language model to emit GarmentCode programs that are then reposed by simulation (about 50\,s).
Their shape comes from the pattern rather than from the image, and is limited by the pattern vocabulary.
Our method generates arbitrary-topology garment geometry from the image in one forward pass, with no templates or test-time optimization.
\vspace{-0.5em}

\subsection{Garment Animation}
Garments are usually animated by binding a canonical garment to the body and deforming it.
Linear blend skinning (LBS)~\cite{magnenat1988lbs} is fast but rigid and tears or stiffens loose cloth; learned retargeting methods~\cite{patel2020tailornet,santesteban2022snug} predict pose-dependent deformation, with DrapeNet~\cite{deluigi2023drapenet} draping arbitrary tops and bottoms from a shared latent space, and HOOD~\cite{grigorev2023hood} simulates garment dynamics with a hierarchical graph network.
All of them need a canonical garment, a binding and a per-frame deformation step; our method generates the garment directly on the target body and is more accurate than the aforementioned methods (\cref{tab:animation}).

\subsection{Garment--Body Training Data}
Learning garments conditioned on a body requires data that pairs each garment with the body it was worn on.
Single-surface methods train on real scans such as THuman2.0~\cite{yu2021function4d}, and 4D-DRESS~\cite{wang20244ddress} and CloSe~\cite{antic2024close} are garment-labeled benchmarks, but their bodies are registrations fitted to the clothed scan.
Template and pattern methods trained on synthetic garments such as CLOTH3D~\cite{bertiche2020cloth3d} inherit their categories.
BEDLAM2.0~\cite{tesch2025bedlam2} renders SMPL-X bodies in simulated clothing to train body estimators~\cite{patel2025camerahmr,shen2024gvhmr,wang2025prompthmr}; we use them for the clothing instead, with the body it was simulated on as the condition.
Earlier work closed the synthetic-to-real gap on the garment side~\cite{jiang2020bcnet,moon2022clothwild}; we close it on the body side, taking the body as is from off-the-shelf HMR systems trained on real images~\cite{shen2024gvhmr,wang2025prompthmr,sam3dbody2025}.

%% file: sec/3_method.tex
\vspace{-0.5em}
\section{Method}
\label{sec:method}

Given an RGB crop $I$ of a person, an SMPL-X body $B$~\cite{pavlakos2019smplx} and, optionally, a garment mask $M$, \method generates the mesh $G$ of the selected garment, or of the whole outfit, around $B$ (\cref{fig:pipeline}).
At test time, $B$ is estimated by an off-the-shelf HMR system such as GVHMR~\cite{shen2024gvhmr}, PromptHMR~\cite{wang2025prompthmr} or SAM 3D Body~\cite{sam3dbody2025}, and $M$ by SAM~\cite{kirillov2023sam} or FastSAM~\cite{zhao2023fastsam}; the full-outfit mask generates the whole outfit at once.
We train two models with the same architecture: the \emph{masked model} takes $M$ and generates the selected garment, and the \emph{main model} takes only $I$ and $B$ and generates the whole outfit.
They differ only in the mask input and in their training pairs: the masked model is trained on mismatched image--body pairs (\cref{sec:method_pairs}), the main model on aligned pairs only.
We first describe the body-conditioned generator (\cref{sec:method_gen}) and how training it on mismatched image--body pairs lets one model both reconstruct and retarget garments (\cref{sec:method_transfer}).
We then introduce a fine-tuning stage that teaches the generator to avoid the body (\cref{sec:penetration}), and finally describe inference (\cref{sec:method_inference}).

\subsection{Body-Conditioned Garment Generation}
\label{sec:method_gen}

We generate garments in the sparse structured latent space of TRELLIS.2~\cite{xiang2025trellis2}, whose encoders and decoders we keep frozen.
This space represents a shape at two levels: a dense \emph{structure} latent $z_\text{ss}\in\mathbb{R}^{8\times16^3}$, which the structure decoder maps to a $64^3$ occupancy grid, and a \emph{structured latent} $z_\text{slat}\in\mathbb{R}^{N\times32}$ on the $N$ occupied cells of a $32^3$ grid, which the shape decoder maps to an O-Voxel~\cite{xiang2025trellis2} at $512^3$ resolution: each active voxel stores a vertex position near the voxel and flags for which neighbouring vertices it connects to.
We train generators for both levels, conditioned on the body and the image.
We first define an image-aligned space shared by the garment, body and image (\cref{sec:method_space}), then describe how the body and image condition the generators (\cref{sec:method_cond}) and how the two stages are trained (\cref{sec:method_fm}).

\subsubsection{Image-Aligned Garment Space}
\label{sec:method_space}

We represent body and garments in a normalized cube $[-\tfrac12,\tfrac12]^3$ whose front face coincides with the $512\times512$ input crop.
A vertex is placed at the crop pixel $(u,v)$ onto which it projects in the camera, and its third coordinate is vertex's depth relative to the pelvis in crop pixels.
Viewed orthographically from the front, the cube therefore reproduces the crop.
Since the crop window, the camera intrinsics and the pelvis depth are known, the mapping is invertible in closed form.

\subsubsection{Body and Garment Conditioning}
\label{sec:method_cond}

In this space, the body is given to the generator as orthographic normal maps from four viewpoints: front, back, left and right.
The crop, the garment mask and the four body renders are encoded separately by a frozen encoder~\cite{simeoni2025dinov3}, and their tokens are concatenated into a single context that both generators attend to through cross-attention in every block; the main model is without the mask.

\subsubsection{Two-Stage Flow Matching}
\label{sec:method_fm}

Both generators are 1.3B-parameter diffusion transformers (DiTs)~\cite{peebles2023dit} trained with flow matching~\cite{lipman2023flow,liu2023rectified}.
For a target latent $z$, noise $\epsilon\sim\mathcal{N}(0,I)$ and a time $\tau\in[0,1]$, the noisy latent $x_\tau=(1-\tau)\,z+\tau\,\epsilon$ follows a straight path, and the DiT $v_\theta$ regresses its velocity,
\begin{equation}
  \mathcal{L}_\text{FM}=\mathbb{E}_{z,\epsilon,\tau}\,\big\|v_\theta(x_\tau,\tau,y)-(\epsilon-z)\big\|^2,
\end{equation}
with condition $y=(I,M,B)$ ($y=(I,B)$ for the main model), dropped in a fraction of the steps to enable classifier-free guidance.
Stage~1 generates $z_\text{ss}$, whose decoded occupancy is max-pooled to the $32^3$ grid, giving the $N$ active cells that the garment occupies.
Stage~2 generates $z_\text{slat}$ on these cells, and the shape decoder produces the garment mesh.

\subsection{Generation and Retargeting in a Single Model}
\label{sec:method_transfer}

We first describe the training data and then how pairing each body with images from other frames lets the same model both reconstruct and retarget garments.

\subsubsection{Training Data}
\label{sec:method_data}

We train on BEDLAM2.0~\cite{tesch2025bedlam2}, which provides photorealistic renders of SMPL-X bodies and 10,592 unique, cloth-simulated garment combinations, obtained through size, texture and garment-template variations.
The garments are simulated on each animated body.
Real scans~\cite{yu2021function4d,wang20244ddress,antic2024close} provide neither: their bodies are registrations fitted to the clothed surface that partly pierce the garment, and their clothing is one fused surface without the layers hidden underneath.
We use 507,940 frames in which the person is fully visible, holding out separate sequences for validation, and encode each ground-truth garment, placed in the image-aligned cube of its frame, with the frozen encoders into the Stage~1 and Stage~2 targets $z_\text{ss}$ and $z_\text{slat}$.

\begin{figure}[t]
  \centering
  \includegraphics[width=0.9\columnwidth]{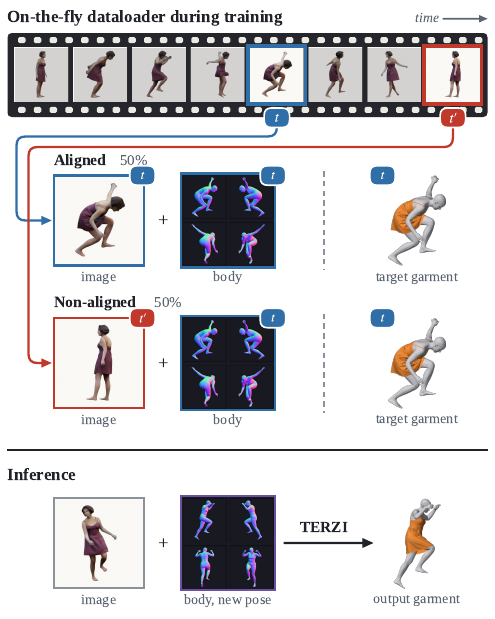}
  \caption{\textbf{Data mix.} Each training example takes its target garment and body renders from a frame $t$ of a BEDLAM2.0 sequence.
  The image comes either from the same frame (\emph{aligned}) or from another frame $t'$ of the same sequence (\emph{non-aligned}).
  This makes the model take only the garment from the image, not the pose, so the garment can be transferred to different poses.}
  \label{fig:train_mix}
\end{figure}

\begin{figure*}[t]
  \centering
  \includegraphics[width=0.9\linewidth]{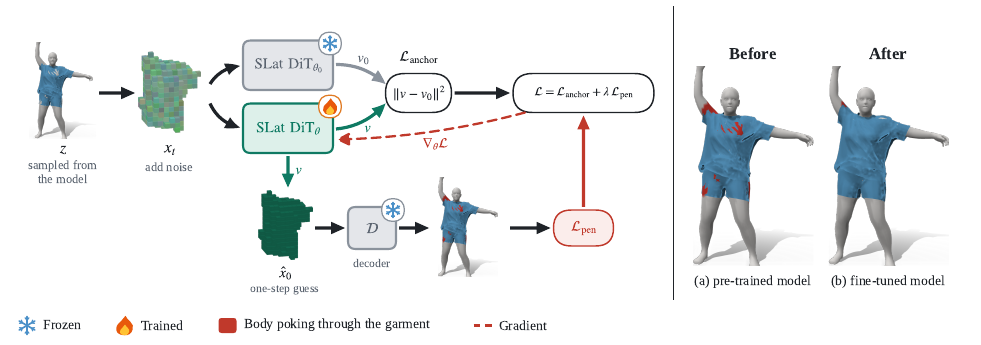}
  \caption{\textbf{Penetration fine-tuning.}
  \emph{Left:} a garment $z$ sampled from the model is noised to $x_\tau$ and passed to the trained DiT ($\theta$) and a frozen copy ($\theta_0$).
  The anchor loss $\mathcal{L}_\text{anchor}=\|v-v_0\|^2$ keeps the trained prediction close to the pre-trained one.
  The trained velocity $v$ also gives a one-step estimate $\hat{x}_0$, which the frozen decoder turns into a garment on the body; $\mathcal{L}_\text{pen}$ penalizes the body poking through it.
  DiT is updated, with $\mathcal{L}=\mathcal{L}_\text{anchor}+\lambda\,\mathcal{L}_\text{pen}$.
  \emph{Right:} the sample before and after fine-tuning (red: body poking through the garment).}
  \label{fig:ftpen}
\end{figure*}

\subsubsection{Mismatched Image--Body Pairs}
\label{sec:method_pairs}

If the image always shows the body in the same pose, the model can copy the garment's pose from the image instead of taking it from the body.
We therefore break this correspondence when training the masked model (\cref{fig:train_mix}): the crop comes from the same frame as the body 50\% of the time, from a different frame of the same sequence 40\% of the time, and from the A-pose, thus a different frame, in the remaining 10\%.
The garment and the body always come from the same frame, so the target garment follows the body, not the image.
The model thus learns to read the garment's identity and appearance from the image alone and to take its pose and shape from the conditioning body.
\subsubsection{Reconstruction and Retargeting}
\label{sec:method_retarget}

At inference, the same model serves two settings without retraining.
With the body estimated from the input image, it reconstructs the garment aligned with the image.
With a different target body, it \emph{retargets} the garment: it generates the garment read from the source image directly on the new body or pose.
Retargeting therefore needs no canonical garment, no skinning weights and no simulation; applied frame by frame along a body sequence (\cref{tab:animation}). This also allows us to generate an A-pose garment for downstream use.

\subsection{Collision-Aware Fine-Tuning}
\label{sec:penetration}

We fine-tune the DiT to reduce the body-garment collisions. Since flow-matching loss supervises latents, not surfaces, nothing prevents the decoded garment from passing through the body.
Moreover, the GT simulated training garments themselves penetrate their bodies slightly (1.82\%).
Rather than correcting collisions after generation, we fine-tune the surface generator with a penalty on the decoded geometry (\cref{fig:ftpen}): a penetration loss pushes the garment out of the body, while an anchor to the pre-trained model keeps it unchanged elsewhere.

\subsubsection{Fine-Tuning Objective}
\label{sec:method_ft_obj}

We fine-tune only the Stage~2 DiT $v_\theta$ (SLat), initialized from the trained model, of which we keep a frozen copy $v_{\theta_0}$.
The training data are the model's own samples: for each training garment, a latent $z$ generated by $v_{\theta_0}$ on the active cells of its garment, so the fine-tuning needs no extra data or labels.
At a random time $\tau$, we noise $z$ to $x_\tau=(1-\tau)\,z+\tau\,\epsilon$ on the same cells, exactly as in Stage~2 training, and pass $x_\tau$ to both DiTs, which predict the velocities $v=v_\theta(x_\tau,\tau,y)$ and $v_0=v_{\theta_0}(x_\tau,\tau,y)$.
The penalty alone would let the garment drift or shrink. The anchor loss $\mathcal{L}_\text{anchor}=\|v-v_0\|^2$ keeps the fine-tuned model close to the pre-trained one. The objective combines the anchor with the penalty,
\begin{equation}
  \mathcal{L}=\mathcal{L}_\text{anchor}+\lambda\,\mathcal{L}_\text{pen},
\end{equation}
and is backpropagated into the DiT; the decoder and the frozen copy stay fixed.
The frozen model is an exact optimum of this term, so without the penalty the fine-tuning is a no-op and every change comes from the penalty.

\subsubsection{Penetration Loss}
\label{sec:method_pen}

From $v$ and $x_\tau$ we take the one-step flow estimate of the clean latent, $\hat{x}_0=x_\tau-\tau\,v$, and the frozen shape decoder $\mathcal{D}$ turns it into the vertices $V=\mathcal{D}(\hat{x}_0)$ of a garment on the conditioning body.
Each vertex is the position the decoder predicts for its O-Voxel cell, a smooth function of the latent, so gradients flow through the decoder into the DiT.
For each vertex we compute its signed distance $s_i$ to the body (negative inside) and penalize vertices that lie inside the body or closer to it than a clearance $\delta$,
\begin{equation}
  \mathcal{L}_\text{pen}=\frac{1}{|V|}\sum_{i=1}^{|V|}q(s_i),\quad q(s)=\min(r,c)^2+2c\,\max(0,r-c),
\end{equation}
where $r=\max(0,\delta-s)$ is the violation, \ie how far a vertex falls short of the clearance, and $c$ is the largest displacement the decoder can apply to a single vertex, which stays within two voxels of its cell.
The function $q$ is thus zero for vertices that keep the clearance, quadratic for violations up to $c$, and linear beyond.

\subsubsection{Effect of Fine-Tuning}
\label{sec:method_ft_effect}

The training framework itself already keeps penetration low: the masked model is trained on crops that in half of the samples show a different pose than the body (\cref{sec:method_transfer}), so it cannot copy the garment's pose from the image and instead "wraps" the garment it reads from the image around the conditioning body. 
Even without fine-tuning, it penetrates only 1.93\% on 4D-DRESS, against 5.50\% for the main model, trained on aligned pairs, and 7.60\% for a model trained without the body.
The fine-tuning lowers penetration further for both models, to 0.58\% and 1.65\% on 4D-DRESS, below the 3.14\% of the GT scans against their own fitted bodies, and to 0.89\% and 1.49\% on BEDLAM2.0, below the 1.82\% of the simulated garments the model is trained on.
Beyond penetration, it improves almost every accuracy metric, NC on all benchmarks and CD and F@1 on 4D-DRESS, and reduces holes and detached seams by removing the decoder's double-sheet artifacts.
The details are given in the supplementary material.

\subsection{Inference}
\label{sec:method_inference}

At inference, each stage integrates the guided velocity with 12 Euler steps, and holes smaller than 10\,cm in the decoded mesh are filled.
We then apply the inverse of the image-aligned mapping to the generated garment and to the body, which returns both to metric camera coordinates.
Garments generated in separate runs using different masks have significantly fewer collisions with the body, but not necessarily with one another.
For fast inference, we distill each 12-step guided stage into a one-step student~\cite{liu2023rectified} that regresses the straight path from the noise to the teacher's sample, with guidance baked in.
With CUDA graphs and compiled networks, the student generates all garments of an image in about 0.1\,s.
For video, sharing the noise across frames and blending the intermediate latents of neighboring frames during sampling reduces flicker without retraining (details in the supplementary material).

%% file: sec/4_experiments.tex
\section{Experiments}
\label{sec:experiments}

We first describe the evaluation setup (\cref{sec:exp_setup}) and compare our method with prior work on scan benchmarks (\cref{sec:exp_recon}) and in the wild (\cref{sec:exp_wild}).
We then show per-garment generation and retargeting (\cref{sec:exp_pergarment,sec:exp_transfer}).

\subsection{Evaluation Setup}
\label{sec:exp_setup}

We report Chamfer distance (CD, cm), normal consistency (NC) and F-score at 1\,cm (F@1) on the garment surface, after aligning each prediction to the scan with scaled ICP.
Since our method and some baselines output separate garment layers while others output a single fused surface, we evaluate under two protocols: \emph{GT label transfer} labels the visible surface of every prediction by the nearest scan face, and \emph{Segmentation} evaluates each method's own garment output, segmenting fused surfaces first.
Further details on the benchmarks, and protocols are given in the supplementary material.
Throughout the paper, the best result in each table is in \textbf{bold} and the second best is \underline{underlined}.

\begin{table}[t]
  \centering
  \caption{\textbf{Generation comparison with prior work on 4D-DRESS~\cite{wang20244ddress}} (64 frames; ours on the GVHMR~\cite{shen2024gvhmr} body).
  \emph{Segmentation}: garments are evaluated after segmenting the prediction; \emph{GT label transfer}: garment labels are transferred from the ground-truth scan.
  $^\dagger$\,Closed source; numbers taken from the paper.
  $^*$\,After segmentation.}
  \label{tab:main}
  \setlength{\tabcolsep}{3.5pt}
  \resizebox{\linewidth}{!}{%
  \begin{tabular}{@{}lcccccc@{}}
    \toprule
    & \multicolumn{3}{c}{Segmentation protocol} & \multicolumn{3}{c}{GT label transfer protocol} \\
    \cmidrule(lr){2-4}\cmidrule(l){5-7}
    Method & CD\,$\downarrow$ & NC\,$\uparrow$ & F@1\,$\uparrow$ & CD\,$\downarrow$ & NC\,$\uparrow$ & F@1\,$\uparrow$ \\
    \midrule
    ReMu~\cite{vuran2025remu}              & 3.306 & .732 & .260 & 2.848 & .744 & .273 \\
    TRELLIS.2$^*$~\cite{xiang2025trellis2} & 1.879 & .772 & .413 & 1.846 & .771 & .418 \\
    HumanNOVA$^*$~\cite{hu2026humannova}   & 2.024 & .805 & .382 & 1.958 & .803 & .389 \\
    SiTH$^*$~\cite{ho2024sith}             & 2.755 & .768 & .263 & 2.599 & .768 & .273 \\
    ISP~\cite{li2023isp}                   & 2.882 & .783 & .282 & 2.695 & .770 & .296 \\
    ChatGarment~\cite{bian2025chatgarment} & 2.707 & .786 & .284 & 2.452 & .788 & .307 \\
    DeClotH$^{\dagger}$~\cite{nam2025declot} & 3.902 & -- & -- & -- & -- & -- \\
    Dress-1-to-3$^\dagger$~\cite{li2025dress1to3}   & 2.441 & -- & -- & -- & -- & -- \\
    \midrule
    \method (ours)              & \textbf{1.484} & \textbf{.813} & \textbf{.481} & \textbf{1.477} & \textbf{.816} & \textbf{.485} \\
    \method (ours, masked)      & \underline{1.528} & \underline{.811} & \underline{.465} & \underline{1.503} & \underline{.815} & \underline{.476} \\
    \bottomrule
  \end{tabular}}
\end{table}

\begin{figure*}[t]
  \centering
  \includegraphics[width=0.9\linewidth]{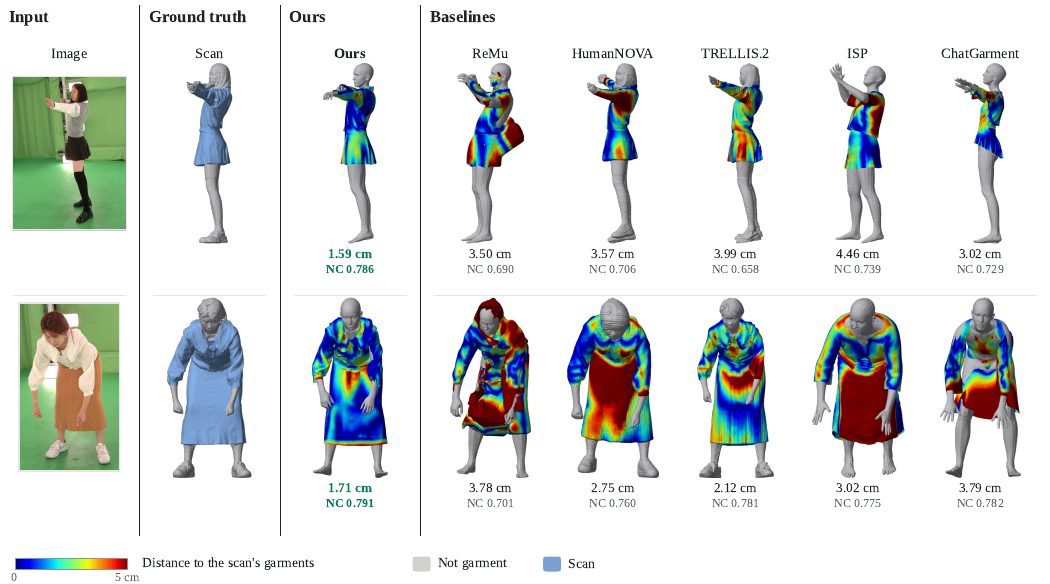}
  \caption{\textbf{Qualitative comparison on 4D-DRESS~\cite{wang20244ddress}.}
  Garments are colored by distance to the scan's garments (0--5\,cm); the body is shown in gray. The protocol here is the segmentation protocol, and per-sample CD and NC are given below each result.}
  \label{fig:qualitative}
\end{figure*}

\subsection{Comparison with Prior Work}
\label{sec:exp_recon}
\cref{tab:main} compares \method against decomposed~\cite{li2023isp,nam2025declot,li2025dress1to3,vuran2025remu,bian2025chatgarment} and single-surface~\cite{xiang2025trellis2,hu2026humannova,ho2024sith} methods; the single-surface methods are segmented into body and garment for evaluation.
Our method is best on all metrics, and \cref{fig:qualitative} shows it stays accurate on hard poses and loose garments where the baselines fail.
Conditioning on the body, and taking only the garment from the image (instead of the pose), is what makes this possible: the same model trained without the body, generating body and outfit jointly from the image, is less accurate (1.584 vs.\ 1.484\,cm CD on 4D-DRESS), lets 7.60\% of the garment penetrate its own body instead of 1.65\%, and can generate anatomically impossible bodies whose errors its garments inherit.
Nor does this depend on a particular body estimate: with each of six recent HMR methods, our method stays ahead of every baseline on all three benchmarks.
These ablations are given in the supplementary material, which also shows that our method degrades gracefully when the person is partly out of frame.
Our method is also fast: it generates the garments in a feed-forward manner in 2.2\,s (0.1\,s with the distilled student), whereas ReMu and ISP fit the garment to each image at test time and take 10.4\,min and 46\,min per image.

\begin{table}[t]
  \centering
  \caption{\textbf{In-the-wild garment normals} on RICH~\cite{huang2022rich} and 3DPW~\cite{vonmarcard2018recovering} against Sapiens2~\cite{khirodkar2026sapiens2} pseudo-ground-truth normals.
  $^*$\,After segmentation.}
  \label{tab:wild}
  \setlength{\tabcolsep}{3.5pt}
  \resizebox{\linewidth}{!}{%
  \begin{tabular}{@{}lcccccccc@{}}
    \toprule
    & \multicolumn{4}{c}{RICH} & \multicolumn{4}{c}{3DPW} \\
    \cmidrule(lr){2-5}\cmidrule(l){6-9}
    Method & Mean\,$\downarrow$ & Med.\,$\downarrow$ & $<$$11.25^\circ$\,$\uparrow$ & $<$$22.5^\circ$\,$\uparrow$ & Mean\,$\downarrow$ & Med.\,$\downarrow$ & $<$$11.25^\circ$\,$\uparrow$ & $<$$22.5^\circ$\,$\uparrow$ \\
    \midrule
    TRELLIS.2$^*$~\cite{xiang2025trellis2} & 29.41 & 25.50 & 17.0 & 46.2 & 28.85 & 24.65 & 15.0 & 45.1 \\
    HumanNOVA$^*$~\cite{hu2026humannova}   & \underline{20.21} & \underline{17.79} & \underline{25.8} & \underline{65.9} & \underline{20.50} & \underline{17.75} & \underline{26.6} & \underline{66.0} \\
    SiTH$^*$~\cite{ho2024sith}             & 27.21 & 23.85 & 18.1 & 50.0 & 24.48 & 21.03 & 21.4 & 55.4 \\
    ChatGarment~\cite{bian2025chatgarment} & 28.52 & 23.82 & 17.7 & 48.0 & 27.93 & 23.50 & 18.5 & 48.6 \\
    \midrule
    \method (ours)                         & \textbf{19.05} & \textbf{14.57} & \textbf{38.0} & \textbf{72.3} & \textbf{19.55} & \textbf{14.94} & \textbf{36.6} & \textbf{70.4} \\
    \bottomrule
  \end{tabular}}
  \vspace{-1.5em}
\end{table}

\begin{figure}[t]
  \centering
  \includegraphics[width=0.95\columnwidth]{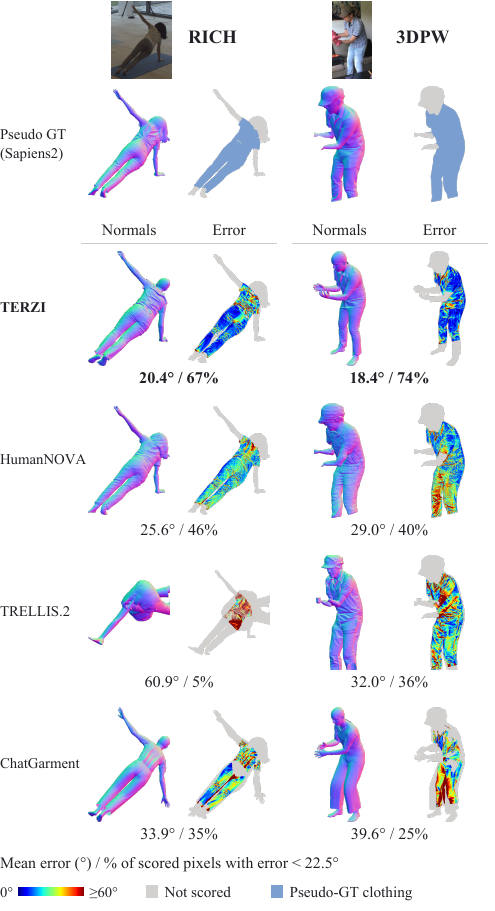}
  \caption{\textbf{In-the-wild garment normals} on RICH~\cite{huang2022rich} (left) and 3DPW~\cite{vonmarcard2018recovering} (right).
  For each example, the first row shows each method's output rendered as normals in its own image space, next to the Sapiens2 pseudo-ground-truth (GT) normals; the second row shows the angular error to the pseudo ground truth on the pixels that are clothing in the pseudo ground truth and garment in the prediction. Numbers: mean error and share of pixels below $22.5^\circ$.}
  \label{fig:wild}
  \vspace{-1.5em}
\end{figure}

\subsection{In-the-Wild Evaluation}
\label{sec:exp_wild}

Real photographs have no 3D ground truth, so we compare garment surface normals against pseudo-ground-truth normals and clothing labels predicted by Sapiens2~\cite{khirodkar2026sapiens2}, on 94 RICH~\cite{huang2022rich} and 128 3DPW~\cite{vonmarcard2018recovering} test frames.
Our method has the most accurate garment normals on both datasets (\cref{tab:wild}), with the largest gap at fine tolerance: $38.0\%$ and $36.6\%$ of our pixels are within $11.25^\circ$, against $25.8\%$ and $26.6\%$ for the strongest baseline, HumanNOVA (\cref{fig:wild}).

\subsection{Per-Garment Generation}
\label{sec:exp_pergarment}

Because the garment is selected by the input mask, the same model generates individual garments or the complete outfit.
Garments generated in separate runs sit on the same conditioning body, so they assemble into a layered outfit with only the light layer resolve, and an open outer jacket is generated as its own layer over the upper garment.
Details are given in the supplementary material.

\subsection{Garment Retargeting}
\label{sec:exp_transfer}

Transfer to new poses is evaluated quantitatively as animation below (\cref{tab:animation}). Details and visuals are given in the supplementary material.

\begin{table}[t]
  \centering
  \caption{\textbf{Garment animation on 4D-DRESS~\cite{wang20244ddress} and BEDLAM2.0~\cite{tesch2025bedlam2}} validation split. 
  The two-stage route binds the ground-truth (GT) A-pose garment of the same outfit and moves it to the target pose with LBS, DrapeNet~\cite{deluigi2023drapenet} or HOOD~\cite{grigorev2023hood}; \method direct instead generates the garment on the target body in one pass. HOOD is run only on the inner garments, and failures are dropped in baselines, for a fair comparison.}
  \label{tab:animation}
  \setlength{\tabcolsep}{3.5pt}
  \resizebox{\linewidth}{!}{%
  \begin{tabular}{@{}lcccccc@{}}
    \toprule
    & \multicolumn{3}{c}{GT garment + retargeting} & \multicolumn{3}{c}{\method direct} \\
    \cmidrule(lr){2-4}\cmidrule(l){5-7}
    Retargeting & CD\,$\downarrow$ & NC\,$\uparrow$ & F@1\,$\uparrow$ & CD\,$\downarrow$ & NC\,$\uparrow$ & F@1\,$\uparrow$ \\
    \midrule
    \multicolumn{7}{@{}l}{\textit{4D-DRESS}} \\
    \quad LBS (per garment)                     & 1.742 & \textbf{.815} & .430 & \textbf{1.681} & .806 & \textbf{.446} \\
    \quad DrapeNet~\cite{deluigi2023drapenet}   & 1.823 & .804 & .410 & \textbf{1.681} & \textbf{.806} & \textbf{.446} \\
    \quad HOOD~\cite{grigorev2023hood}, static  & 1.492 & .849 & .440 & \textbf{1.218} & \textbf{.857} & \textbf{.532} \\
    \quad HOOD~\cite{grigorev2023hood}, motion  & 1.503 & .846 & .435 & \textbf{1.202} & \textbf{.860} & \textbf{.536} \\
    \midrule
    \multicolumn{7}{@{}l}{\textit{BEDLAM2.0}} \\
    \quad LBS (per garment)                     & 0.992 & .862 & .739 & \textbf{0.631} & \textbf{.891} & \textbf{.889} \\
    \quad DrapeNet~\cite{deluigi2023drapenet}   & 1.152 & .847 & .633 & \textbf{0.616} & \textbf{.886} & \textbf{.901} \\
    \quad HOOD~\cite{grigorev2023hood}, static  & 1.088 & .852 & .694 & \textbf{0.621} & \textbf{.893} & \textbf{.893} \\
    \quad HOOD~\cite{grigorev2023hood}, motion  & 1.169 & .845 & .666 & \textbf{0.617} & \textbf{.894} & \textbf{.895} \\
    \bottomrule
  \end{tabular}}
  \vspace{-1.5em}
\end{table}

A common way to dress a body in a new pose is to obtain the garment once in a canonical pose and animate it.
We compare this two-stage route with generating the garment directly on the target body (\cref{tab:animation}; the details are in the supplementary material).
In the two-stage route, an A-pose garment, either our generation or the ground-truth garment of the same outfit, is bound to the body through the skinning weights of its nearest body vertices, smoothed over the garment mesh, and moved to the target pose by LBS, by the learned draping model DrapeNet~\cite{deluigi2023drapenet}, or by the neural simulator HOOD~\cite{grigorev2023hood}.
Each method must produce the subject's garments in a target pose and is scored against the garment actually worn in the target frame.
On 4D-DRESS, we use the 64 frames of \cref{tab:main} with GVHMR target bodies, and our input image comes from the same take at least 20 frames away from the target frame; on BEDLAM2.0, we use 188 validation frames.
HOOD either interpolates the body from the A-pose to the target pose (\emph{static}) or follows the real motion leading to the target frame (\emph{motion}).
Generating the garment directly on the target body gives the lowest CD with every retargeting method on both datasets, ahead of the ground-truth garment and of our own A-pose garment.
On 4D-DRESS direct generation is ahead of LBS, DrapeNet and HOOD.
DrapeNet takes about 120\,s and HOOD 3.5--7\,min per frame, on top of generating the A-pose garments.

%% file: sec/5_conclusion.tex
\vspace{-0.8em}
\section{Conclusion}
\label{sec:conclusion}

\method recovers separate, image-aligned body and garment meshes from a single image by generating the garments around a given body.
It brings three components together.
Feed-forward, body-conditioned generation without test-time optimization.
Training on mismatched image--body pairs lets the same model retarget the garment read from the image to a new pose or body, without skinning or simulation.
Collision-aware fine-tuning teaches the generator to keep the garment outside the body, no collision correction is needed at inference.
Trained only on simulated BEDLAM2.0 garments, the model is the most accurate method on real scans (4D-DRESS, CloSe) and in-the-wild photographs (RICH, 3DPW) while being the fastest method compared.
Treating the body as a condition rather than an output turns garment reconstruction and retargeting into one generation problem; the same formulation could extend to other layers worn on the body, and to other descriptions of the garment than an image.

\vspace{-1.0em}

\paragraph{Limitations and future work.}
Output mesh quality is bounded by the sparse shape decoder of TRELLIS.2, and our method inherits its limitations, including small holes and detached seams in the garment surface.
Our method also predicts each frame independently: a sequence is reconstructed or animated frame by frame. Blending the latents of neighboring frames at inference time reduces flicker, but learning temporal consistency and cloth dynamics during training remains future work.
Finally, while showing synthetic data alone can support state-of-the-art garment reconstruction, higher-quality simulated garments offer a promising way to expand the range of garments it can generate.

\paragraph{LLM usage statement.}
Claude Opus 5.5 were used for code creation and lightly edit \cref{sec:related}, \cref{sec:method}, and \cref{sec:experiments} for clarity.
No LLMs were used for the research ideas, hypotheses, claims or conclusions, or for the design and analysis of the experiments.
The authors take full responsibility for all content.

%% file: sec/X_suppl.tex
\clearpage
\setcounter{page}{1}
\maketitlesupplementary

\renewcommand{\topfraction}{0.9}
\renewcommand{\dbltopfraction}{0.9}
\renewcommand{\bottomfraction}{0.8}
\renewcommand{\textfraction}{0.07}
\renewcommand{\floatpagefraction}{0.7}
\renewcommand{\dblfloatpagefraction}{0.7}
\setcounter{topnumber}{3}
\setcounter{dbltopnumber}{2}
\setcounter{totalnumber}{5}

\newcommand{\suppentry}[2]{\noindent\hyperref[#1]{\makebox[2em][l]{\ref*{#1}}#2}\dotfill\pageref*{#1}\par}
\newcommand{\suppsubentry}[2]{\noindent\hspace*{2em}\hyperref[#1]{\makebox[2.5em][l]{\ref*{#1}}#2}\dotfill\pageref*{#1}\par}

\section*{Contents}
{\small
\setlength{\parskip}{0.15em}
  \suppentry{sec:supp_training}{Training Details}
  \suppsubentry{sec:supp_space}{Image-Aligned Garment Space}
  \suppsubentry{sec:supp_schedules}{Data and Schedules}
  \suppentry{sec:supp_collision}{Collision-Aware Fine-Tuning}
  \suppentry{sec:supp_layer}{Layer Resolve}
  \suppentry{sec:supp_eval}{Evaluation Protocol}
  \suppentry{sec:supp_benchmarks}{Additional Benchmarks}
  \suppentry{sec:supp_wild}{In-the-Wild Evaluation}
  \suppentry{sec:runtime}{Run Time}
  \suppentry{sec:supp_ablation}{Ablations}
  \suppentry{sec:supp_occlusion}{Robustness to Truncation}
  \suppentry{sec:supp_pergarment}{Per-Garment Generation}
  \suppsubentry{sec:supp_mask_robust}{Robustness to the Garment Mask}
  \suppentry{sec:supp_layered}{Layered Garment Evaluation}
  \suppentry{sec:supp_transfer}{Garment Retargeting}
  \suppentry{sec:supp_anim}{Garment Animation}
  \suppsubentry{sec:supp_anim_protocol}{Protocol}
  \suppsubentry{sec:supp_anim_methods}{Methods}
  \suppsubentry{sec:supp_anim_runtime}{Run Time and Paired Comparison}
  \suppsubentry{sec:supp_anim_smooth}{Smoothed vs.\ Nearest-Vertex Binding}
  \suppentry{sec:supp_significance}{Statistical Significance}
  \suppentry{sec:supp_animation}{Qualitative Animation Results}
  \suppentry{sec:supp_video}{Video Reconstruction}
  \suppentry{sec:supp_texture}{Texturing}
}
\vspace{1em}

\section{Training Details}
\label{sec:supp_training}

\subsection{Image-Aligned Garment Space}
\label{sec:supp_space}

\subsubsection{Crop}
Given an image with intrinsics $K=(f_x,f_y,c_x,c_y)$ and an estimated body, we take a square window of side $s$ (pixels) centred at $(\bar u,\bar v)$ on the projected body, padded by a factor of $1.2$, and resize it to $512\times512$.
The crop therefore has the focal length $f_\text{crop}=512\,f_x/s$ in crop pixels.
The crop is chosen once, from the image, and both the conditioning body and the target garments are mapped into the $[-\tfrac12,\tfrac12]^3$ cube in which the encoders and DiTs operate.

\subsubsection{From Camera to Cube}
Let $P=(X,Y,Z)$ be a vertex in camera coordinates and $Z_p$ the depth of the body's pelvis.
Its crop coordinates are
\begin{align}
  u &= \tfrac{512}{s}\Big(f_x\tfrac{X}{Z}+c_x-\bar u\Big)+256, \\
  v &= \tfrac{512}{s}\Big(f_y\tfrac{Y}{Z}+c_y-\bar v\Big)+256, \\
  w &= (Z-Z_p)\,\frac{f_\text{crop}}{Z_p},
\end{align}
that is, the perspective projection into the crop for $u,v$ and a pelvis-anchored depth in crop pixels for $w$.
The cube coordinates are $(x,y,z)=\big((u-256)/512,\,(v-256)/512,\,0.6\,w/512\big)$, so the front face spans the crop and the pelvis lies at $z=0$.
Every vertex lies on the pixel where it appears in the image, and an orthographic view of the cube along $z$ reproduces the crop.
The fixed depth gain of $0.6$ gives the cube a depth half-extent of $512/1.2\approx427$ crop pixels, enough for poses that reach toward the camera; without it, $4\%$ of training samples were clipped in depth.
The gain changes neither the alignment of the front face with the crop nor the invertibility of the mapping.
Geometry outside the crop falls outside the cube, so a partially visible person yields a partially reconstructed garment.

\subsubsection{From Cube Back to Camera}
The mapping is inverted vertex by vertex.
From $(x,y,z)$ we recover $u=512x+256$, $v=512y+256$ and $w=512\,z/0.6$, the image pixel $u_\text{img}=\bar u-\tfrac s2+u\,\tfrac{s}{512}$ (likewise $v_\text{img}$), and the depth $Z=Z_p\,(1+w/f_\text{crop})$, then back-project with the pinhole model, $X=(u_\text{img}-c_x)\,Z/f_x$ and $Y=(v_\text{img}-c_y)\,Z/f_y$.
The garment and the body use the same crop, intrinsics and $Z_p$, so both return to the same metric camera frame, where they can be compared with ground truth or placed back into the scene; all evaluations and renderings use these metric meshes.

\subsubsection{Body Renders and Condition Tokens}
The body is rendered in the cube with an orthographic camera at $512\times512$ from four yaw angles about the cube's vertical axis: front ($0^\circ$), back ($180^\circ$), left ($-90^\circ$) and right ($+90^\circ$).
We use flat per-face normals, two-sided towards the viewer, coloured as $\tfrac12(n+1)$ on a black background, the same convention as the RGB condition.
The front render is pixel-aligned with the crop.
In the side views the horizontal axis is the pelvis-relative depth in crop pixels.
The renders undo the depth gain, so the side views show the body with its true proportions.
A frozen DINOv3~\cite{simeoni2025dinov3} encodes the RGB crop and the garment mask into $1{,}029$ tokens each and every body view into $261$ tokens.
Their concatenation, with tokens of dimension $1{,}024$, is the cross-attention context of every block of both DiTs: $2\times1{,}029+4\times261=3{,}102$ tokens for the masked model, and $1{,}029+4\times261=2{,}073$ for the main model, which has no mask.

\subsection{Data and Schedules}
\label{sec:supp_schedules}

\subsubsection{BEDLAM2.0}
\label{sec:supp_bedlam}
BEDLAM2.0~\cite{tesch2025bedlam2} is a synthetic video dataset of SMPL-X~\cite{pavlakos2019smplx} bodies rendered in Unreal Engine~5 with ground-truth bodies and cameras.
Its 10{,}592 cloth-simulated garment combinations (\cref{sec:method_data}) were designed by professional clothing designers in the CLO fashion design software and range from single dresses to multi-layer outfits in which each garment, \eg trousers, shirt and jacket, is a separate mesh.
Graded garments come in ten sizes (XS--6XL), and each avatar wears the size that matches its body mass index; each outfit has 6--28 texture variations, and failed simulations are removed after visual inspection.

\subsubsection{Training Data}
Both teachers are trained on the same corpus of single-person frames from this dataset.
The 16,057 single-person sequences with clothing simulation, sampled at 40 frames each, give 642,280 posed samples.
Keeping only frames in which the reprojected person lies fully inside the image leaves 510,271 samples from 14,718 sequences, split into 507,940 for training and 2,331 for validation, with no sequence or outfit shared between the two.
Each sample consists of a garment latent (the outfit), a structure latent, the RGB crop and the four body renders.
\Cref{tab:training} summarizes the schedules below.

\subsubsection{Main Model}
\label{sec:supp_main}
The main model is conditioned on the RGB crop and the four body views and generates the whole outfit.
The structure (SS) and the surface (SLat) stages are initialized from the pretrained flow matching models.

\subsubsection{Masked Model}
\label{sec:supp_masked}
The masked model is additionally conditioned on a garment mask and is initialized from the main model.
It is trained on the same 507,940 training samples, but each sample serves several targets: the whole outfit or any single garment, giving 988,924 single-garment targets (outfits have 1 garment in 77k samples, 2 in 301k, 3 in 119k, and more in the rest), plus A-pose targets for 325,125 samples.
For robustness, the RGB crop comes from the same frame 50\% of the time and from a different frame of the same sequence 40\% of the time, and from the A-pose frame in the remaining 10\%.
For both stages we keep the checkpoint with the best validation loss.
The teachers evaluated throughout the paper, \method and \method (masked), are these two models after a short penetration fine-tuning of the surface stage (\cref{sec:supp_collision}); the structure stage is unchanged.
They sample with 12 + 12 Euler steps (structure + surface) and classifier-free guidance 7.5.

\begin{table}[!htbp]
  \centering
  \caption{\textbf{Training schedules.} Batch size is GPUs $\times$ samples per GPU; passes are over the 507,940 training samples.}
  \label{tab:training}
  \setlength{\tabcolsep}{3pt}
  \resizebox{\linewidth}{!}{%
  \begin{tabular}{@{}llccr@{}}
    \toprule
    Model & Stage & Batch & LR & Samples seen \\
    \midrule
    Main   & SS   & $16\times8=128$  & $5\cdot10^{-4}$ & 3.2\,M ($\approx$6 passes) \\
           & SLat & $16\times12=192$ & $10^{-4}$       & 16.7\,M ($\approx$33 passes) \\
    \midrule
    Masked & SS   & $4\times32=128$  & $5\cdot10^{-4}$ & 2.9\,M \\
           & SLat & $4\times48=192$  & $2\cdot10^{-5}$ & 6.1\,M ($\approx$12 passes) \\
    \midrule
    Pen.\ FT & SLat & $1\times8=8$   & $10^{-5}$       & 800 \\
    \bottomrule
  \end{tabular}}
\end{table}

\subsubsection{Distillation}
\label{sec:supp_distill_train}
We distill each teacher stage (structure and surface) into a one-step student, initialized from the teacher and trained on a separate pool of 48,665 training outfits with cached condition tokens; the masked model trains on the whole outfit 20\% of the time and otherwise on one random garment with its mask.
Every example draws fresh noise $\epsilon$ and runs a full 12-step guided teacher sample $\hat{x}_0$ (\cref{sec:supp_masked}).
The student is first trained with reflow~\cite{liu2023rectified} along the straight path between the two,
\begin{equation}
  \mathcal{L}_\text{reflow}=\mathbb{E}_{\tau}\,\big\|v_\psi(x_\tau,\tau,y)-(\epsilon-\hat{x}_0)\big\|^2,\quad x_\tau=(1-\tau)\,\hat{x}_0+\tau\,\epsilon,
\end{equation}
which straightens its trajectories, and then distilled into a single step from the reflowed model's own samples, $\mathcal{L}_\text{distill}=\|v_\phi(\epsilon,1,y)-(\epsilon-\hat{x}_0)\|^2$, so that $x_0=\epsilon-v_\phi(\epsilon,1,y)$.
Guidance is baked into the targets, so the student runs without guidance.
Training uses 1 GPU, 4 examples per update and learning rate $10^{-5}$, for 6,000 + 6,000 updates (reflow + distillation) for the surface stage and 3,000 + 3,000 for the structure stage.

\section{Collision-Aware Fine-Tuning}
\label{sec:supp_collision}

\subsection{Data}
The training pairs are samples of the frozen teacher itself: for each training outfit, the generated latent $z$ together with the noise $\epsilon$ that produced it, for 1,520 outfits (main model) and 601 outfits (masked model).
The body $B$ is the watertight SMPL-X body of each sample.
Anchoring to these samples rather than to the ground-truth garments keeps the model from imitating the slight penetration of the simulated garments (\cref{sec:method_ft_obj}).

\subsection{Training Step}
\Cref{tab:collision_settings} lists all settings.
We draw $\tau\sim\mathcal{U}(0.02,0.98)$ and noise $z$ as in Stage~2 training, with a noise floor of $\sigma_{\min}=10^{-5}$.
The penalty is applied for $\tau\le0.6$.
This threshold is not tuned: it is the lower bound of the guidance interval $\tau\in[0.6,1]$ of the pre-trained sampler~\cite{xiang2025trellis2}, below which the sampler integrates the unguided conditional velocity that the penalty trains.
The decoded vertex positions are an affine map of a sigmoid of the decoder output, so they are differentiable in the latent; only the connectivity is discrete.
Signed distances to the body are computed with an exact BVH query.
They are measured in crop pixels of the image-aligned space (\cref{sec:supp_space}), about 3.79\,mm on average, so the clearance $\delta=1.5$ crop pixels is $\approx$5.7\,mm and the linear threshold $c=2$ crop pixels $\approx$7.6\,mm.
The threshold $c$ is the furthest the decoder can move a vertex: each vertex is placed within $[-0.5,1.5]$ voxels of its cell, \ie at most two voxels, which is 2 crop pixels in $x$ and $y$ at the $512^3$ output resolution (3.33 in $z$ because of the depth gain).
Deeper violations therefore cannot be fixed by moving vertices, and letting them dominate pushed earlier attempts into deleting surface.

\begin{table}[!htbp]
  \centering
  \caption{\textbf{Collision-aware fine-tuning settings} of the teachers used throughout the paper.}
  \label{tab:collision_settings}
  \setlength{\tabcolsep}{3.5pt}
  \resizebox{\linewidth}{!}{%
  \begin{tabular}{@{}ll@{}}
    \toprule
    Setting & Value \\
    \midrule
    Initialization & pre-trained SLat DiT \\
    Trained / frozen & SLat DiT / SS DiT, decoder, DINOv3 \\
    Anchor & frozen copy of the initial SLat DiT, weight 1 \\
    Penetration weight $\lambda$ & 30 \\
    Penalty noise range & $\tau\le0.6$ (= lower bound of the sampler's guidance interval) \\
    Clearance $\delta$ & 1.5 crop px $\approx$ 5.7\,mm \\
    Linear threshold $c$ & 2 crop px $\approx$ 7.6\,mm \\
    Condition dropout (CFG) & 10\% \\
    Optimizer, learning rate & AdamW, $10^{-5}$ \\
    Batch & 8 examples per step (gradient accumulation 8), 1 GPU \\
    Checkpoint & after 100 steps (800 examples) \\
    Max.\ active cells per sample & 1{,}200 \\
    Validation score & penetration + 0.2 $\times$ drift from the frozen surface \\
    \bottomrule
  \end{tabular}}
\end{table}

\subsection{Penetration in the Ground Truth}
\label{sec:supp_gt_pen}
Even the ground-truth data are not penetration-free (\cref{tab:gt_penetration}).
On 4D-DRESS, 3.14\% of the scanned garment surface lies inside the released SMPL-X fit on average, and 1.55\% more than 2\,mm deep; on the BEDLAM2.0 validation split, 1.82\% of the simulated garments lies inside the ground-truth body.
After fine-tuning, both of our models penetrate their conditioning body less than the 4D-DRESS scans penetrate their own fitted body, by mean, median and deep share, although they are conditioned on an estimated rather than a fitted body.
A penetration of 0 is therefore not a realistic target.
The masked model penetrates less than the main model, for the reason discussed in \cref{sec:supp_abl_collision}.

\begin{table}[!htbp]
  \centering
  \caption{\textbf{Body penetration in the ground truth and in our outputs.}
  Percentage of the garment surface inside the body; deep: more than 2\,mm inside.
  Ground truth: 4D-DRESS scan garments \vs the released SMPL-X fit, and BEDLAM2.0 simulated garments \vs the ground-truth body.
  Ours: fine-tuned teachers, default output (\cref{sec:supp_postproc}) \vs the conditioning body (GVHMR on 4D-DRESS; \cref{tab:collision_ablation}).}
  \label{tab:gt_penetration}
  \setlength{\tabcolsep}{3.5pt}
  \resizebox{\linewidth}{!}{%
  \begin{tabular}{@{}lcccc@{}}
    \toprule
    & Frames & Mean\,(\%) & Median\,(\%) & Deep\,(\%) \\
    \midrule
    \multicolumn{5}{@{}l}{\textit{4D-DRESS}} \\
    Ground truth (scan \vs SMPL-X fit) & 64 & 3.14 & 2.43 & 1.55 \\
    \method                            & 64 & 1.65 & 1.26 & 0.96 \\
    \method (masked)                   & 64 & 0.58 & 0.11 & 0.33 \\
    \midrule
    \multicolumn{5}{@{}l}{\textit{BEDLAM2.0 val}} \\
    Ground truth (simulation \vs body) & 200 & 1.82 & 0.73 & 1.22 \\
    \bottomrule
  \end{tabular}}
\end{table}

\subsection{Results}
The clearance $\delta$ and the checkpoint are selected on the held-out BEDLAM2.0 validation split, with the validation score of \cref{tab:collision_settings}; no test benchmark is used for this choice.
\Cref{tab:collision_ablation} shows the same trend on 4D-DRESS (64 outfits, GVHMR bodies).
The clearance $\delta$ is the main lever: penetration falls steadily as $\delta$ grows from 0.13 to 2.0 crop pixels, for both models.
Accuracy does not suffer and even improves slightly (NC $+0.016$ and F@1 $+0.008$ for the main model at $\delta=1.5$), because the fine-tune also removes the double-sheet artifacts of the decoder.
On 4D-DRESS, a larger clearance would also tear the surface: the floating fragments of the main model grow from about 0.7--1.3\% without fine-tuning to about 2\% at $\delta=1.0$--1.5, and to 5.4--5.9\% at $\delta=2.0$, where 44--51 of the 64 outfits have more than 2\% floating surface.
Hole metrics do not catch this, so we also monitor fragments.
For the masked model, fragments level off at about 2\% from $\delta=1.5$ onwards.
Training longer brings no further gain (step 200 to 1200 for the main model, 200 to 400 for the masked model), consistent with the selected checkpoint at step 100 (800 examples).
A weaker anchor (weight 0.3) or a higher learning rate ($3\cdot10^{-5}$) is clearly worse, with less penetration reduction and more drift.
In further runs with SAM 3D Body~\cite{sam3dbody2025} bodies, not included in the table, multi-step rollouts of the clean estimate (2--12 steps), $\lambda$ of 100 or 300 (more drift and shrinking), a wider clip $c$, and applying the penalty at higher noise levels ($\tau$ up to 0.8 or 1.0) brought no gain.

\begin{table}[!tbp]
  \centering
  \caption{\textbf{Ablation: collision-aware fine-tuning} on 4D-DRESS (64 outfits, GVHMR bodies, raw outputs).
  Unless a step is given, each row is the checkpoint at step 100 (800 examples), with $\lambda=30$ and the anchor.
  $\delta$: clearance in crop pixels.
  Pen.: garment surface inside the body (\%), in parentheses deeper than 2\,mm.
  $^\dagger$Tears the surface (see text).
  These are raw outputs without hole filling, under the evaluation used during development, so absolute values differ from those of \cref{tab:abl_collision} (\eg CD 1.452 here \vs 1.484 there for the same model); only comparisons within this table are meaningful.}
  \label{tab:collision_ablation}
  \setlength{\tabcolsep}{3.5pt}
  \resizebox{\linewidth}{!}{%
  \begin{tabular}{@{}lcccc@{}}
    \toprule
    Setting & CD\,$\downarrow$ & NC\,$\uparrow$ & F@1\,$\uparrow$ & Pen.\,(\%)\,$\downarrow$ \\
    \midrule
    \multicolumn{5}{@{}l}{\textit{\method}} \\
    No fine-tune                     & 1.453 & 0.803 & 0.488 & 5.47 (3.18) \\
    $\delta=0.13$, step 200          & 1.457 & 0.808 & 0.489 & 4.31 (2.46) \\
    $\delta=0.13$, step 1200         & 1.454 & 0.805 & 0.489 & 4.52 (2.54) \\
    $\delta=0.5$                     & 1.455 & 0.811 & 0.489 & 3.79 (2.16) \\
    $\delta=1.0$                     & 1.458 & 0.814 & 0.492 & 2.68 (1.49) \\
    $\delta=1.5$ (used)              & 1.452 & 0.819 & 0.496 & 1.63 (0.95) \\
    $\delta=2.0^\dagger$             & 1.457 & 0.823 & 0.494 & 0.90 (0.54) \\
    Anchor weight 0.3 (instead of 1) & 1.454 & 0.812 & 0.490 & 3.91 (2.20) \\
    LR $3\cdot10^{-5}$ (instead of $10^{-5}$) & 1.454 & 0.803 & 0.489 & 4.60 (2.62) \\
    \midrule
    \multicolumn{5}{@{}l}{\textit{\method (masked)}} \\
    No fine-tune                     & 1.510 & 0.805 & 0.472 & 1.88 (0.94) \\
    $\delta=0.13$, step 200          & 1.506 & 0.810 & 0.473 & 1.20 (0.54) \\
    $\delta=0.13$, step 400          & 1.510 & 0.808 & 0.473 & 1.20 (0.56) \\
    $\delta=0.5$                     & 1.503 & 0.810 & 0.476 & 1.06 (0.54) \\
    $\delta=1.0$                     & 1.497 & 0.813 & 0.479 & 0.69 (0.40) \\
    $\delta=1.5$ (used)              & 1.478 & 0.818 & 0.486 & 0.53 (0.30) \\
    $\delta=2.0$                     & 1.467 & 0.823 & 0.490 & 0.39 (0.23) \\
    Anchor weight 0.3 (instead of 1) & 1.499 & 0.811 & 0.477 & 1.10 (0.55) \\
    LR $3\cdot10^{-5}$ (instead of $10^{-5}$) & 1.495 & 0.810 & 0.477 & 2.00 (0.97) \\
    \bottomrule
  \end{tabular}}
\end{table}

\section{Layer Resolve}
\label{sec:supp_layer}

Collision-aware fine-tuning keeps generated garment outside the body, so no body--garment collision correction is needed; all reported numbers or qualitative visualizations use none.
When garments are generated in separate runs with different masks and then combined, an inner garment (\eg a shirt) can poke through an outer one (\eg a jacket), since each is generated without knowing the others.
A light layer resolve fixes this: going from the inside out, every vertex that lies below the layers beneath it, measured as height above its nearest body point along the body normal, is lifted 3\,mm above them, and the push fades out smoothly over the surrounding garment.
Loose drape is kept, and the resolve takes about 40\,ms per outfit.

\section{Evaluation Protocol}
\label{sec:supp_eval}

\subsection{Benchmarks}
On 4D-DRESS~\cite{wang20244ddress}, following DeClotH~\cite{nam2025declot}, we use 64 frames, with one take per outfit (Inner and Outer outfits) and one randomly selected frame per take, seen from the frontal capture camera.
DeClotH~\cite{nam2025declot} and Dress-1-to-3~\cite{li2025dress1to3} are closed source, so their numbers in \cref{tab:main} are taken from their papers rather than recomputed under our protocol, and are given for reference only.
On CloSe~\cite{antic2024close} we use the 146 scans of the CloSe-Di test split.
Each scan is rendered from a frontal camera placed 4\,m in front of the subject at chest height (30$^\circ$ field of view, $1024\times1024$, the scan's vertex colors, white background).
On these two scan benchmarks, the ground truth is the garment surface of the scan, \ie all faces labeled as clothing (4D-DRESS: upper, lower and outer; CloSe: all classes except body, hair, hat and shoes).
On BEDLAM2.0~\cite{tesch2025bedlam2} we use 200 frames from held-out validation sequences, rendered with the dataset's own camera.
Here the ground truth is the simulated clothing, reduced to its visible surface so that it matches a scan.
All methods receive the same input image crop.
Our model is conditioned on an estimated body only, from GVHMR~\cite{shen2024gvhmr} on 4D-DRESS and from PromptHMR~\cite{wang2025prompthmr} on CloSe and BEDLAM2.0; no ground-truth pose or shape is used.
The masked model receives the ground-truth whole-outfit mask, rendered from the garment labels; Grounded SAM~\cite{ren2024groundedsam} masks give no significant difference on 4D-DRESS (\cref{sec:supp_mask_robust}).
All recent estimators we tested perform similarly, and with each of them our method stays ahead of every baseline (\cref{tab:abl_body}); in the comparisons with prior work we use GVHMR on 4D-DRESS and PromptHMR on CloSe, where video-based estimators are not applicable; on the held-out BEDLAM2.0 validation split, these two are also the most accurate estimators.

\subsection{Metrics}
Following SiTH~\cite{ho2024sith} and 4D-DRESS~\cite{wang20244ddress}, we sample 100k points uniformly, each with the normal of its face, on the predicted and on the ground-truth garment surface.
The Chamfer distance (CD, cm) is the mean of the two directed nearest-neighbor distances: prediction to ground truth (P2S, accuracy), which penalizes spurious or misplaced geometry, and ground truth to prediction (S2P, completeness), which penalizes holes and missing parts.
Normal consistency (NC) is the mean absolute cosine between the normal of each point and that of its nearest neighbor in the other set, averaged over both directions; the absolute value makes it independent of normal orientation.
The F-score at 1\,cm (F@1) is the harmonic mean of precision, the fraction of predicted points within 1\,cm of the ground truth, and recall, the fraction of ground-truth points within 1\,cm of the prediction.
Unlike CD, it is not dominated by a few distant outliers.
All metrics are computed per frame and averaged over the benchmark.
Sampling seeds are fixed per frame, so all methods are evaluated against identical ground-truth samples.

\subsection{Mesh-Quality Metrics}
\label{sec:supp_mesh_metrics}
Besides CD, NC and F@1, we report two properties of the generated garment mesh itself, which are identical under both protocols.

\emph{Penetration} (pen., \%) is the percentage of the garment surface that lies inside our own conditioning body.
We sample 100k points on the garment and count those with a negative signed distance to the body mesh.
It is defined only for our models, which output garment and body as separate meshes in the same frame.
Lower is better, down to the level of the ground truth itself (\cref{sec:supp_gt_pen}).

\emph{Holes} counts the boundary loops of the raw garment mesh with a perimeter of at least 10\,cm, including the natural openings of the garment (neckline, sleeves, waist, hems); increases thus indicate additional holes.

\subsection{Alignment}
Single-view reconstructions are defined only up to scale and global pose, and each method outputs in its own coordinate frame.
Following the evaluation of SiTH~\cite{ho2024sith}, we therefore align each prediction to the scan with ICP~\cite{besl1992icp} with uniform scale.
The complete prediction, \ie the garments together with the method's own body or full surface, is aligned to the complete scan, and the garments inherit this transformation.
Errors in garment placement relative to the body are therefore not compensated by a per-garment alignment.
ICP is initialized by matching the principal axes of the prediction and the scan.
On CloSe and BEDLAM2.0, the sign ambiguity of the principal axes is resolved by running ICP from each of the four sign combinations and keeping the solution with the lowest surface distance.

\subsection{Two Protocols}
The methods differ in output structure: our method, ReMu~\cite{vuran2025remu}, ISP~\cite{li2023isp} and ChatGarment~\cite{bian2025chatgarment} produce separate garment layers, whereas HumanNOVA~\cite{hu2026humannova}, TRELLIS.2~\cite{xiang2025trellis2} and SiTH~\cite{ho2024sith} produce a single fused surface of body and clothing.
We therefore report two complementary protocols that differ only in which part of the prediction is treated as the garment.

\emph{Segmentation} evaluates the garment output of each method, \ie the garment mesh that would be used downstream, including its hidden parts.
For layered methods these are their garment layers.
For fused-surface methods we segment the garment from the surface following the multi-view recipe of ReMu~\cite{vuran2025remu}: we render 60 views of the colored mesh, segment each view in 2D, back-project the labels and assign each face by majority vote.
As the 2D segmenter we use the union of SCHP human parsing~\cite{li2020schp} and Grounded-SAM~\cite{ren2024groundedsam}.
Because hidden surfaces are included, this protocol also penalizes incorrect inner geometry.

\emph{GT label transfer} is segmentation-free.
A real scan only contains the externally visible surface, so we first extract the visible shell of the aligned prediction.
We render it orthographically ($1536\times1536$) from 252 directions (36 azimuths $\times$ 7 elevations) and keep every face visible in at least one view.
This removes hidden layers, inner garment surfaces and the body beneath the clothing.
Each remaining face then takes the label of its nearest ground-truth scan triangle, and the faces labeled as clothing form the evaluated garment.
The garment is thus defined by the same rule for every method, independent of its output structure or its own segmentation, and the protocol measures how well the visible clothed surface matches the scan. It guards against segmentation errors: methods that output a single fused surface must be segmented before evaluation, and any error in this step would be charged to the reconstruction itself, so this protocol gives a fair comparison.

\subsection{Post-Processing}
\label{sec:supp_postproc}
Baseline outputs are evaluated as produced.
For our method we fill holes with a perimeter below 10\,cm and apply no smoothing; small holes of this kind can mostly be rectified with such standard mesh post-processing~\cite{xiang2025trellis2}.
This hole filling is applied by default for visual quality only (\cref{tab:postproc}).
Under the segmentation protocol it changes every metric by at most .003. Under GT label transfer, it slightly worsens CD (by up to .052\,cm) and F@1 (by up to .021) on 4D-DRESS and CloSe; all reported numbers of our models therefore include this small penalty.

\begin{table}[!htbp]
  \centering
  \caption{\textbf{Effect of post-processing} on the metrics of our method.
  Difference between the default output (hole filling below 10\,cm) and the output without any post-processing (default minus none), under both protocols.
  Protocols as in \cref{tab:main}.}
  \label{tab:postproc}
  \setlength{\tabcolsep}{3.5pt}
  \resizebox{\linewidth}{!}{%
  \begin{tabular}{@{}lcccccc@{}}
    \toprule
    & \multicolumn{3}{c}{Segmentation} & \multicolumn{3}{c}{GT label transfer} \\
    \cmidrule(lr){2-4}\cmidrule(l){5-7}
    Model & $\Delta$CD & $\Delta$NC & $\Delta$F@1 & $\Delta$CD & $\Delta$NC & $\Delta$F@1 \\
    \midrule
    \multicolumn{7}{@{}l}{\textit{4D-DRESS}} \\
    \method          & $+$.002 & $-$.000 & $-$.001 & $+$.023 & $-$.003 & $-$.011 \\   
    \method (masked) & $+$.003 & $-$.000 & $+$.000 & $+$.029 & $-$.003 & $-$.012 \\   
    \midrule
    \multicolumn{7}{@{}l}{\textit{CloSe}} \\
    \method          & $+$.000 & $+$.000 & $-$.000 & $+$.047 & $-$.003 & $-$.020 \\   
    \method (masked) & $-$.001 & $-$.000 & $+$.000 & $+$.052 & $-$.004 & $-$.021 \\   
    \bottomrule
  \end{tabular}}
\end{table}

\section{Additional Benchmarks}
\label{sec:supp_benchmarks}

\begin{figure*}[!tp]
  \centering
  \includegraphics[width=\linewidth]{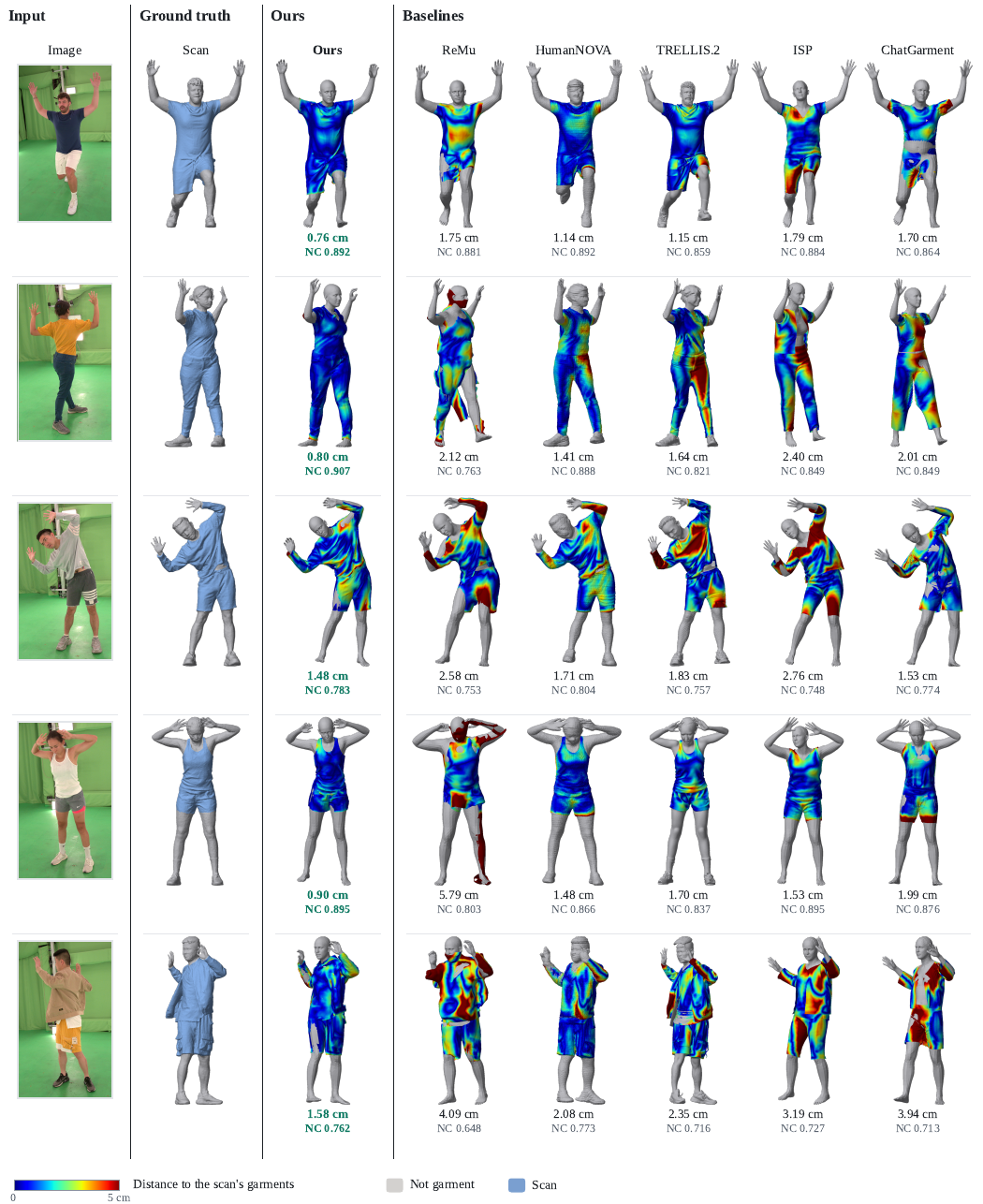}
  \caption{\textbf{Additional qualitative comparison on 4D-DRESS~\cite{wang20244ddress}}, as in \cref{fig:qualitative}.
  Garments are colored by distance to the scan's garments (0--5\,cm); the body and non-garment parts are shown in gray.
  Per-sample CD and NC are given below each result.}
  \label{fig:qual_4dd_more}
\end{figure*}

\begin{figure*}[!tp]
  \centering
  \includegraphics[width=\linewidth]{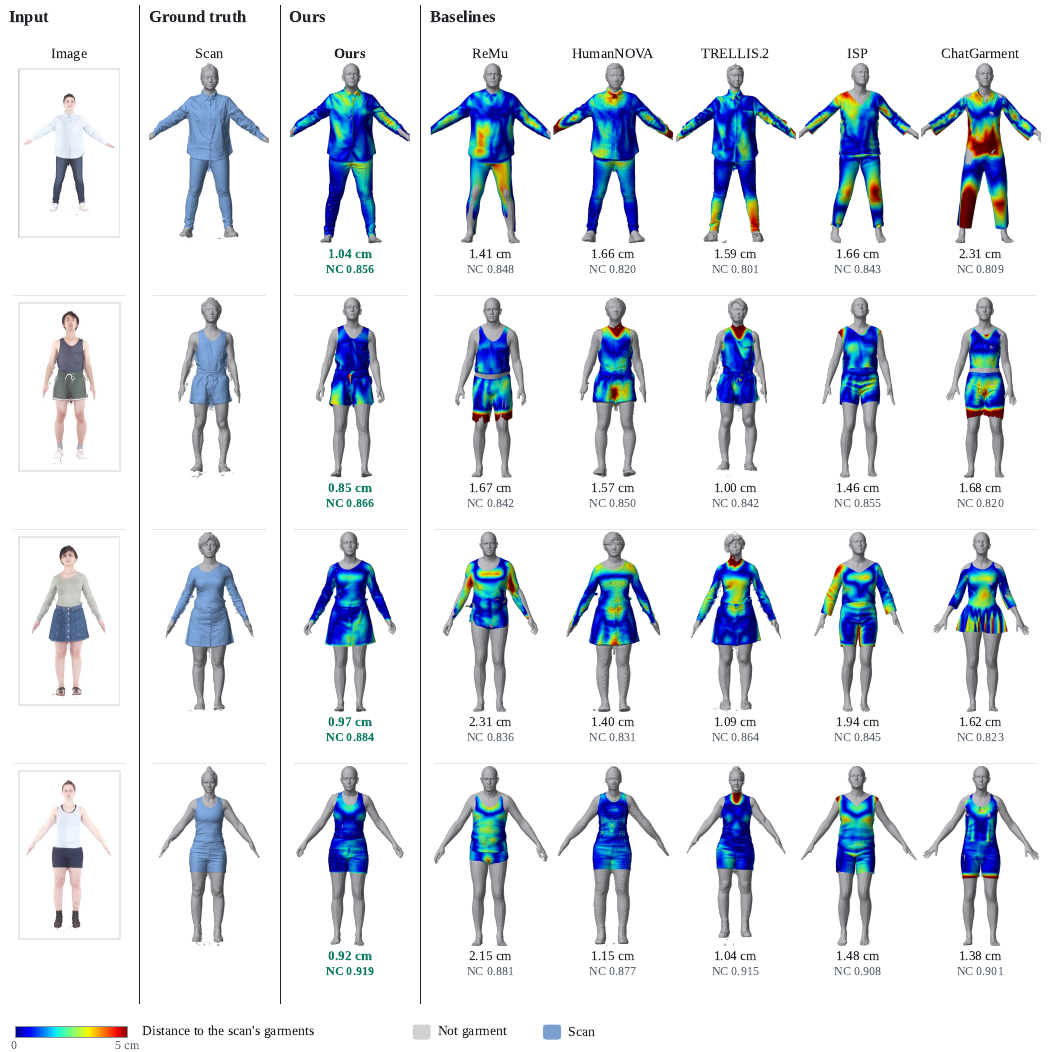}
  \caption{\textbf{Qualitative comparison on CloSe~\cite{antic2024close}}, as in \cref{fig:qualitative}.
  Garments are colored by distance to the scan's garments (0--5\,cm); the body and non-garment parts are shown in gray.
  Per-sample CD and NC are given below each result.}
  \label{fig:qual_close}
\end{figure*}

\Cref{tab:close} reports results on CloSe~\cite{antic2024close}.
\Cref{fig:qual_4dd_more,fig:qual_close} give further qualitative comparisons with all baselines on 4D-DRESS and CloSe.

\begin{table}[!htbp]
  \centering
  \caption{\textbf{Results on CloSe~\cite{antic2024close}} (146 frontal frames; ours on the PromptHMR~\cite{wang2025prompthmr} body).
  Protocols as in \cref{tab:main}.
  $^*$\,After segmentation.
  $^\dagger$\,Failed on some frames and averaged over the rest: ReMu on 3 frames (143 evaluated), ISP on 1 (145 evaluated).}
  \label{tab:close}
  \setlength{\tabcolsep}{3.5pt}
  \resizebox{\linewidth}{!}{%
  \begin{tabular}{@{}lcccccc@{}}
    \toprule
    & \multicolumn{3}{c}{Segmentation} & \multicolumn{3}{c}{GT label transfer} \\
    \cmidrule(lr){2-4}\cmidrule(l){5-7}
    Method & CD\,$\downarrow$ & NC\,$\uparrow$ & F@1\,$\uparrow$ & CD\,$\downarrow$ & NC\,$\uparrow$ & F@1\,$\uparrow$ \\
    \midrule
    ReMu$^\dagger$~\cite{vuran2025remu}    & 1.717 & .831 & .406 & 1.554 & .837 & .425 \\ 
    TRELLIS.2$^*$~\cite{xiang2025trellis2} & 1.270 & .818 & \underline{.526} & 1.223 & .819 & .535 \\
    HumanNOVA$^*$~\cite{hu2026humannova}   & 1.303 & \textbf{.837} & \textbf{.534} & 1.220 & .839 & .541 \\
    SiTH$^*$~\cite{ho2024sith}             & 1.618 & \underline{.834} & .402 & 1.567 & .835 & .410 \\
    ISP$^\dagger$~\cite{li2023isp}         & 1.908 & .832 & .375 & 1.721 & .826 & .390 \\
    ChatGarment~\cite{bian2025chatgarment} & 2.170 & .811 & .336 & 1.855 & .817 & .367 \\
    \midrule
    \method (ours)                         & \textbf{1.202} & \textbf{.837} & \textbf{.534} & \textbf{1.148} & \textbf{.843} & \textbf{.553} \\
    \method (ours, masked)                 & \underline{1.228} & \underline{.834} & .523 & \underline{1.167} & \underline{.840} & \underline{.544} \\
    \bottomrule
  \end{tabular}}
\end{table}

\section{In-the-Wild Evaluation}
\label{sec:supp_wild}

This section details the in-the-wild evaluation summarized in \cref{tab:wild} with all baselines (\cref{tab:wild_full}), and qualitative examples are in \cref{fig:wild} of the main paper.

\subsection{Data}
To evaluate on real photographs without 3D ground truth, we compare garment surface normals against pseudo-ground-truth normals and clothing labels predicted by Sapiens2-5B~\cite{khirodkar2026sapiens2}.
We use two datasets with natural clothing: the RICH~\cite{huang2022rich} test split, taking the middle annotated frame of every sequence--camera view (191 candidates), and the 3DPW~\cite{vonmarcard2018recovering} test set, taking one frame per second from its 24 test sequences (885 candidates).
Frames are kept by fixed rules, not by hand: a single person (no second Sapiens2 foreground component of at least 10\% of the person's area, and no other detected person box overlapping the subject's by more than 5\%), at least 380\,px tall (RICH) or 55\% of the image height (3DPW), and fully inside the image; occlusion is allowed.
This yields 94 RICH frames (44 sequences, all 8 test subjects) and 128 3DPW frames.

\subsection{Protocol}
Each method is evaluated in its own image space, without any fitting to the pseudo ground truth.
All image-based methods receive the same person crop and apply their own preprocessing, and each output is rendered with the camera that method reconstructs for, read from its code or outputs.
As TRELLIS.2 outputs objects in a canonical orientation, its view is estimated from its own input silhouette, with near-ties resolved in its favour.
The pseudo-ground-truth normals and clothing labels are warped into each method's image and the normals rotated into its camera frame.
Methods that produce a single fused surface (HumanNOVA, TRELLIS.2, SiTH) are split into garments by the same multi-view parsing pipeline.
We measure the angle between predicted and pseudo-ground-truth normals on the pixels that are clothing in the pseudo ground truth and garment in the prediction, and report its mean, its median, and the share of pixels below $11.25^\circ$, $22.5^\circ$ and $30^\circ$, each averaged per frame.
Our method is conditioned on the SAM 3D Body~\cite{sam3dbody2025} body.

\subsection{Results}
\Cref{tab:wild_full} shows that our garments have the most accurate normals on both datasets on every metric: the lowest mean ($19.1^\circ$ on RICH, $19.6^\circ$ on 3DPW) and median error, and the largest share of pixels under each threshold.
The gap is largest at fine angular tolerances: $38.0\%$ and $36.6\%$ of our pixels are within $11.25^\circ$, against $25.8\%$ and $26.6\%$ for the strongest baseline, HumanNOVA, whose meshes carry striped surface ridges (\cref{fig:wild}).
The other baselines are far behind in mean error.

\begin{table}[!tbp]
  \centering
  \caption{\textbf{In-the-wild garment normals, all baselines}, on RICH~\cite{huang2022rich} (94 frames) and 3DPW~\cite{vonmarcard2018recovering} (128 frames) against Sapiens2~\cite{khirodkar2026sapiens2} pseudo-ground-truth normals, on the pixels that are clothing in the pseudo ground truth and garment in the prediction.
  Angular error in degrees and share of pixels below each threshold (\%), averaged per frame.
  Each method is evaluated in its own image space without any fitting.
  Ours on the SAM 3D Body~\cite{sam3dbody2025} body.
  $^*$\,After segmentation.
  $^\dagger$\,ReMu fails on 4 of the 94 RICH frames (its surface refinement produces no mesh); its RICH numbers are over the remaining 90.}
  \label{tab:wild_full}
  \setlength{\tabcolsep}{4pt}
  \resizebox{\linewidth}{!}{%
  \begin{tabular}{@{}llccccc@{}}
    \toprule
    & Method & Mean\,$\downarrow$ & Med.\,$\downarrow$ & $<$$11.25^\circ$\,$\uparrow$ & $<$$22.5^\circ$\,$\uparrow$ & $<$$30^\circ$\,$\uparrow$ \\
    \midrule
    \multirow{7}{*}{\rotatebox{90}{RICH}}
    & ReMu$^\dagger$~\cite{vuran2025remu}    & 31.63 & 27.13 & 16.4 & 44.1 & 58.8 \\
    & TRELLIS.2$^*$~\cite{xiang2025trellis2} & 29.41 & 25.50 & 17.0 & 46.2 & 61.8 \\
    & HumanNOVA$^*$~\cite{hu2026humannova}   & \underline{20.21} & \underline{17.79} & \underline{25.8} & \underline{65.9} & \underline{82.2} \\
    & SiTH$^*$~\cite{ho2024sith}             & 27.21 & 23.85 & 18.1 & 50.0 & 66.5 \\
    & ISP~\cite{li2023isp}                   & 26.69 & 22.81 & 17.4 & 50.1 & 67.2 \\
    & ChatGarment~\cite{bian2025chatgarment} & 28.52 & 23.82 & 17.7 & 48.0 & 63.7 \\
    \cmidrule{2-7}
    & \method (ours)                         & \textbf{19.05} & \textbf{14.57} & \textbf{38.0} & \textbf{72.3} & \textbf{83.0} \\
    \midrule
    \multirow{7}{*}{\rotatebox{90}{3DPW}}
    & ReMu~\cite{vuran2025remu}              & 25.53 & 21.24 & 21.4 & 54.5 & 70.2 \\
    & TRELLIS.2$^*$~\cite{xiang2025trellis2} & 28.85 & 24.65 & 15.0 & 45.1 & 62.2 \\
    & HumanNOVA$^*$~\cite{hu2026humannova}   & \underline{20.50} & \underline{17.75} & \underline{26.6} & \underline{66.0} & \underline{81.4} \\
    & SiTH$^*$~\cite{ho2024sith}             & 24.48 & 21.03 & 21.4 & 55.4 & 71.7 \\
    & ISP~\cite{li2023isp}                   & 24.50 & 20.63 & 21.2 & 56.1 & 72.7 \\
    & ChatGarment~\cite{bian2025chatgarment} & 27.93 & 23.50 & 18.5 & 48.6 & 64.5 \\
    \cmidrule{2-7}
    & \method (ours)                         & \textbf{19.55} & \textbf{14.94} & \textbf{36.6} & \textbf{70.4} & \textbf{81.8} \\
    \bottomrule
  \end{tabular}}
\end{table}

\section{Run Time}
\label{sec:runtime}

\subsection{Our Method}
\Cref{tab:timing} breaks down the time to generate all garments of one image on a single H200.
In the teacher, the two DiTs take most of the time, because every guided sampling step (\cref{sec:supp_masked}) evaluates the network twice.
The distilled student replaces each chain by a single step without guidance.
Because the student runs each DiT exactly once on a fixed-size input, we capture both DiTs as CUDA graphs and compile their bodies.
We also compile the DINOv3 encoder and the SS decoder, and run the sparse shape decoder as CUDA graphs bucketed by the number of active cells.
This removes the per-kernel launch overhead that dominates a single sampling step (\emph{compiled} columns); the same compilation brings the teacher from 2.2\,s to 0.94\,s.

\begin{table}[!htbp]
  \centering
  \caption{\textbf{Run time per image} (ms, median over 16 4D-DRESS frames, one dedicated NVIDIA H200; all garments of the outfit are generated in one pass).
  Teacher: the sampler of \cref{sec:supp_masked}. Student: one step per stage without guidance.
  Eager: plain PyTorch. Compiled: CUDA graphs after a warm-up pass.
  Body estimation is excluded (see text).
  Totals are slightly higher than the sum of the stages because they include the small overhead between stages.}
  \label{tab:timing}
  \setlength{\tabcolsep}{4pt}
  \resizebox{\linewidth}{!}{%
  \begin{tabular}{@{}lrrrr@{}}
    \toprule
    & \multicolumn{2}{c}{Teacher} & \multicolumn{2}{c}{Student} \\
    \cmidrule(lr){2-3}\cmidrule(l){4-5}
    Stage & eager & compiled & eager & compiled \\
    \midrule
    Body normal renders            & 3.0    & 3.1   & 3.0  & 2.6  \\
    DINOv3 conditioning            & 22.1   & 15.5  & 22.7 & 15.2 \\
    SS DiT (sparse structure)      & 1104.6 & 740.4 & 53.0 & 34.0 \\
    SS decoder                     & 32.0   & 4.7   & 33.0 & 4.4  \\
    SLat DiT (structured latent)   & 977.1  & 144.5 & 51.9 & 7.1  \\
    Shape decoder (mesh)           & 37.8   & 21.1  & 35.1 & 25.9 \\
    Hole filling                   & 5.9    & 5.8   & 5.9  & 5.8  \\
    \midrule
    Total                          & 2185   & 942   & 205  & 101  \\
    \bottomrule
  \end{tabular}}
\end{table}
All times exclude body estimation, which takes 49\,ms per image with PromptHMR, 123\,ms per image with SAM~3D~Body and 23\,ms per video frame with GVHMR.

\subsection{Comparison with Prior Work}
Decomposed methods such as ReMu~\cite{vuran2025remu} and ISP~\cite{li2023isp} recover the garment by fitting it to each image at test time: ISP optimizes the latent codes of its implicit sewing patterns against the image evidence, and ReMu optimizes the geometry of each clothing layer it reconstructs from the image.
This iterative per-subject fitting dominates their cost.
On a single NVIDIA H200, ReMu takes 10.4\,min and ISP 46\,min per image, whereas our method generates the garments in 2.2\,s (0.94\,s compiled), roughly $280\times$ and $1{,}300\times$ faster, while being more accurate (\cref{tab:main}).
The distilled student further reduces this to 0.10\,s, about $6{,}000\times$ faster than ReMu and $27{,}000\times$ faster than ISP.
All times are per image, for the median frame, on one GPU with no other jobs running, and exclude body estimation.
Feed-forward single-surface generators are fast as well (TRELLIS.2: 6.1\,s; HumanNOVA: 1.6\,s on the same GPU), but their output must be segmented afterwards, and they are less accurate.
Because optimization runs per image, its cost also scales linearly with the number of frames, which makes per-frame animation (\cref{tab:animation}) impractical for these methods.
The run time of the animation baselines is given in \cref{sec:supp_anim_runtime}.

\section{Ablations}
\label{sec:supp_ablation}

We ablate the body conditioning itself, the body estimator used for it, and the components behind each contribution: the collision-aware training and the distilled student.
Direct retargeting is evaluated against skinning, learned draping and simulation in \cref{tab:animation} and \cref{sec:supp_anim}.

\subsection{Body Conditioning}
To measure what the body input contributes, we train a union teacher on the same data without body conditioning: it generates body and outfit together from the RGB crop alone (\cref{tab:abl_union}).
Removing the body costs 0.100\,cm CD under segmentation on 4D-DRESS and 0.093\,cm on CloSe, and .036 NC on both; under GT label transfer CD rises by 0.015\,cm on 4D-DRESS and 0.072\,cm on CloSe.
Penetration grows from 1.65\% to 7.6\% on 4D-DRESS and from 1.25\% to 6.2\% on CloSe, although the union teacher is measured against the body it generated itself.
Conditioning on the body is thus what keeps the garment outside it, and it also gives more accurate surface orientation and a better fit to the visible garment.

\begin{figure*}[!tp]
  \centering
  \includegraphics[width=\linewidth]{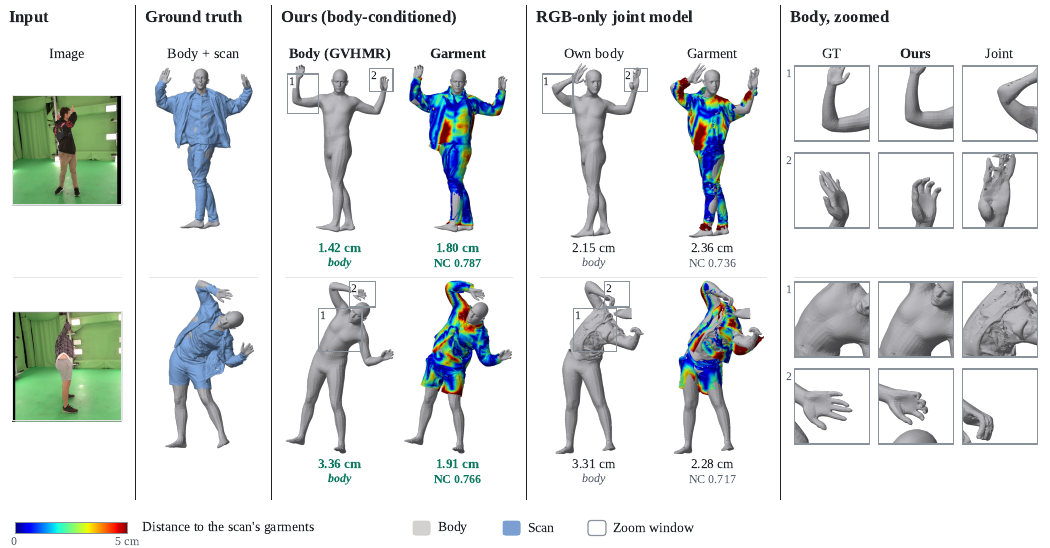}
  \caption{\textbf{A joint model generates implausible bodies, and its garments inherit them.}
  An RGB-only model trained on the same data (the union teacher of \cref{tab:abl_union}) generates body and garment together; \method{} conditions on an estimated parametric body.
  The joint model's bodies can be anatomically impossible: a forearm bent the wrong way with fused fingers (top); garment folds baked into the torso with detached details (bottom).
  Its garments are less accurate in every case, whereas our method dresses a valid parametric body.
  Garments are coloured by their distance to the scan's garments (0--5\,cm); numbers give body Chamfer distance, and garment Chamfer distance with normal consistency.
  Right: the same window on each body.}
  \label{fig:body_error}
\end{figure*}

\Cref{fig:body_error} shows why.
Without a body prior, the union teacher must infer the body from pixels as well, and its bodies can be anatomically impossible: a forearm bent against the elbow, garment folds baked into the torso, or fused fingers.
Because body and garment are generated together, these errors carry over to the garment, which is less accurate in every example even where the generated body itself is close to the scan in Chamfer distance.
Our method instead dresses a valid parametric body: an imperfect estimate shifts the garment, but it cannot make the body, or the garment on it, implausible.

\begin{table}[!tbp]
  \centering
  \caption{\textbf{Ablation: body conditioning.}
  Union teacher: trained without body input, generating body and outfit from the RGB crop alone.
  Ours on the GVHMR~\cite{shen2024gvhmr} body on 4D-DRESS and on the PromptHMR~\cite{wang2025prompthmr} body on CloSe.
  $^*$Measured against the body generated by the union teacher itself.}
  \label{tab:abl_union}
  \setlength{\tabcolsep}{2.5pt}
  \footnotesize
  \begin{tabular}{@{}lccccccc@{}}
    \toprule
    & \multicolumn{3}{c}{Seg.} & \multicolumn{3}{c}{LT} & \\
    \cmidrule(lr){2-4}\cmidrule(lr){5-7}
    Model & CD\,$\downarrow$ & NC\,$\uparrow$ & F@1\,$\uparrow$ & CD\,$\downarrow$ & NC\,$\uparrow$ & F@1\,$\uparrow$ & Pen.\,$\downarrow$ \\
    \midrule
    \multicolumn{8}{@{}l}{\textit{4D-DRESS} (64 frames)} \\
    Union teacher (RGB only) & 1.584 & .777 & .457 & 1.492 & .792 & .478 & 7.6$^*$ \\
    \method                  & \textbf{1.484} & \textbf{.813} & \textbf{.481} & \textbf{1.477} & \textbf{.816} & \textbf{.485} & \textbf{1.65} \\
    \midrule
    \multicolumn{8}{@{}l}{\textit{CloSe} (146 frames)} \\
    Union teacher (RGB only) & 1.295 & .801 & .509 & 1.220 & .819 & .528 & 6.2$^*$ \\
    \method                  & \textbf{1.202} & \textbf{.837} & \textbf{.534} & \textbf{1.148} & \textbf{.843} & \textbf{.553} & \textbf{1.25} \\
    \bottomrule
  \end{tabular}
\end{table}

\subsection{Body Estimator}
Our method is conditioned on a body estimated from the image.
To test how sensitive it is to this estimate, \cref{tab:abl_body} compares, on 4D-DRESS and CloSe, conditioning on the ground-truth body with the estimates of recent video-based (GVHMR~\cite{shen2024gvhmr}, GENMO~\cite{li2025genmo}) and single-image (PromptHMR~\cite{wang2025prompthmr}, SAM 3D Body~\cite{sam3dbody2025}, CameraHMR~\cite{patel2025camerahmr}, Multi-HMR~\cite{baradel2024multihmr}) human mesh recovery methods.
The choice of estimator matters less than the gap to the ground-truth body: under GT label transfer, the CD of the main model varies by at most 0.09\,cm across estimators on either benchmark, whereas the ground-truth body lowers it by a further 0.21\,cm on 4D-DRESS.
On CloSe, where modern estimators are already accurate, the ground-truth body matches the best estimate (PromptHMR) in CD under segmentation and is clearly better than CameraHMR, while PromptHMR keeps a small edge in F@1 and under the masked protocol; the ground-truth body still produces the highest penetration (2.02\%).
With every estimator, the CD of the main model stays below that of every baseline on 4D-DRESS and CloSe (\cref{tab:main,tab:close}).
Penetration depends more on the estimator (1.04--2.14\% for the main model), while the masked model stays at or below 0.96\% with every estimator.

\begin{table*}[!tp]
  \centering
  \caption{\textbf{Ablation: body estimator} for \method and \method (masked) under both evaluation protocols (\cref{sec:supp_eval}).
  Ground truth: the released SMPL-X fit on 4D-DRESS, the released SMPL fit on CloSe, with 28 corrupted parameter sets corrected.
  Input: V, video-based; I, single-image.
  Video-based estimators are not applicable to CloSe, which consists of single frames.
  Seg.: segmentation; LT: GT label transfer.
  Pen.: garment surface inside the conditioning body (\%); it describes the generated mesh and is the same under both protocols.
  $^\dagger$Multi-HMR fails on one 4D-DRESS frame; its row is averaged over the remaining 63.}
  \label{tab:abl_body}
  \setlength{\tabcolsep}{2.8pt}
  \small
  \begin{tabular}{@{}llccccccccccccccc@{}}
    \toprule
    & & \multicolumn{7}{c}{\method} & \multicolumn{7}{c}{\method (masked)} \\
    \cmidrule(lr){3-9}\cmidrule(l){10-16}
    & & \multicolumn{3}{c}{Seg.} & \multicolumn{3}{c}{LT} & & \multicolumn{3}{c}{Seg.} & \multicolumn{3}{c}{LT} & \\
    \cmidrule(lr){3-5}\cmidrule(lr){6-8}\cmidrule(lr){10-12}\cmidrule(lr){13-15}
    Body & Input & CD\,$\downarrow$ & NC\,$\uparrow$ & F@1\,$\uparrow$ & CD\,$\downarrow$ & NC\,$\uparrow$ & F@1\,$\uparrow$ & Pen.\,$\downarrow$ & CD\,$\downarrow$ & NC\,$\uparrow$ & F@1\,$\uparrow$ & CD\,$\downarrow$ & NC\,$\uparrow$ & F@1\,$\uparrow$ & Pen.\,$\downarrow$ \\
    \midrule
    \multicolumn{16}{@{}l}{\textit{4D-DRESS} (64 frames)} \\
    Ground truth                                   & -- & 1.286 & .823 & .555 & 1.270 & .828 & .563 & 0.42 & 1.283 & .819 & .555 & 1.262 & .826 & .562 & 0.32 \\
    GVHMR~\cite{shen2024gvhmr}                     & V  & 1.484 & .813 & .481 & 1.477 & .816 & .485 & 1.65 & 1.528 & .811 & .465 & 1.503 & .815 & .476 & 0.58 \\
    Multi-HMR$^\dagger$~\cite{baradel2024multihmr} & I  & 1.527 & .815 & .455 & 1.488 & .817 & .471 & 2.08 & 1.579 & .813 & .440 & 1.530 & .815 & .460 & 0.52 \\
    PromptHMR~\cite{wang2025prompthmr}             & I  & 1.533 & .814 & .469 & 1.540 & .815 & .477 & 1.16 & 1.601 & .808 & .451 & 1.594 & .811 & .463 & 0.43 \\
    GENMO~\cite{li2025genmo}                       & V  & 1.537 & .811 & .469 & 1.520 & .813 & .474 & 2.14 & 1.625 & .805 & .445 & 1.596 & .809 & .458 & 0.54 \\
    SAM 3D Body~\cite{sam3dbody2025}               & I  & 1.576 & .814 & .449 & 1.542 & .815 & .466 & 2.13 & 1.638 & .809 & .434 & 1.578 & .812 & .456 & 0.96 \\
    CameraHMR~\cite{patel2025camerahmr}            & I  & 1.603 & .809 & .446 & 1.563 & .812 & .462 & 1.38 & 1.684 & .806 & .429 & 1.635 & .809 & .449 & 0.52 \\
    \midrule
    \multicolumn{16}{@{}l}{\textit{CloSe} (146 frames)} \\
    Ground truth                                   & -- & 1.210 & .837 & .523 & 1.149 & .841 & .546 & 2.02 & 1.258 & .833 & .503 & 1.184 & .838 & .531 & 1.00 \\
    PromptHMR~\cite{wang2025prompthmr}             & I  & 1.202 & .837 & .534 & 1.148 & .843 & .553 & 1.25 & 1.228 & .834 & .523 & 1.167 & .840 & .544 & 0.76 \\
    SAM 3D Body~\cite{sam3dbody2025}               & I  & 1.235 & .837 & .526 & 1.192 & .841 & .541 & 1.04 & 1.262 & .833 & .513 & 1.212 & .838 & .531 & 0.62 \\
    CameraHMR~\cite{patel2025camerahmr}            & I  & 1.260 & .831 & .514 & 1.206 & .837 & .532 & 1.53 & 1.289 & .830 & .498 & 1.231 & .836 & .519 & 0.72 \\
    \bottomrule
  \end{tabular}
\end{table*}

\subsection{Collision-Aware Training}
\label{sec:supp_abl_collision}
\Cref{tab:abl_collision} compares the model with and without the differentiable penetration penalty on the decoded garment, without any collision correction at inference.
On both benchmarks, the fine-tuning lowers penetration by 59--74\% and the number of holes by 22--36\%, since it also removes the double-sheet artifacts of the decoder.
It improves NC everywhere and leaves accuracy essentially unchanged otherwise: CD and F@1 improve in every cell except F@1 of the main model on CloSe under segmentation, which changes by .002.
Part of the penetration reduction comes from training on mismatched image--body pairs (\cref{sec:method_pairs}) rather than from the fine-tuning: without any fine-tuning, the masked model already penetrates far less than the main model: 1.93\% \vs 5.50\% on 4D-DRESS and 1.86\% \vs 4.88\% on CloSe.
The fine-tuning still helps the masked model, lowering its penetration by a further 59--70\%, to 0.58\% and 0.76\%.
The penetration metric is defined in \cref{sec:supp_mesh_metrics}, and the fine-tuning recipe and its clearance study are given in \cref{sec:supp_collision}.

\begin{table*}[!tp]
  \centering
  \caption{\textbf{Ablation: collision-aware fine-tuning (FT)} under both evaluation protocols (\cref{sec:supp_eval}).
  No post-hoc collision correction is applied.
  Pen.\ and holes describe the generated mesh and are the same under both protocols; holes per garment (\cref{sec:supp_mesh_metrics}).
  CloSe: ours on the PromptHMR~\cite{wang2025prompthmr} body.}
  \label{tab:abl_collision}
  \setlength{\tabcolsep}{4pt}
  \begin{tabular}{@{}lcccccccc@{}}
    \toprule
    & \multicolumn{3}{c}{Segmentation} & \multicolumn{3}{c}{GT label transfer} & \multicolumn{2}{c}{Mesh} \\
    \cmidrule(lr){2-4}\cmidrule(lr){5-7}\cmidrule(l){8-9}
    Model & CD\,$\downarrow$ & NC\,$\uparrow$ & F@1\,$\uparrow$ & CD\,$\downarrow$ & NC\,$\uparrow$ & F@1\,$\uparrow$ & Pen.\,(\%)\,$\downarrow$ & Holes\,$\downarrow$ \\
    \midrule
    \multicolumn{9}{@{}l}{\textit{4D-DRESS}} \\
    \method, w/o FT          & 1.516 & .786 & .480 & 1.502 & .799 & .481 & 5.50 & 41.2 \\
    \method, w/ FT           & \textbf{1.484} & \textbf{.813} & \textbf{.481} & \textbf{1.477} & \textbf{.816} & \textbf{.485} & \textbf{1.65} & \textbf{26.4} \\
    \method (masked), w/o FT & 1.562 & .791 & .454 & 1.536 & .801 & .461 & 1.93 & 36.4 \\
    \method (masked), w/ FT  & \textbf{1.528} & \textbf{.811} & \textbf{.465} & \textbf{1.503} & \textbf{.815} & \textbf{.476} & \textbf{0.58} & \textbf{25.5} \\
    \midrule
    \multicolumn{9}{@{}l}{\textit{CloSe}} \\
    \method, w/o FT          & 1.207 & .815 & \textbf{.536} & 1.158 & .830 & .550 & 4.88 & 22.2 \\
    \method, w/ FT           & \textbf{1.202} & \textbf{.837} & .534 & \textbf{1.148} & \textbf{.843} & \textbf{.553} & \textbf{1.25} & \textbf{16.0} \\
    \method (masked), w/o FT & 1.253 & .816 & .514 & 1.199 & .828 & .530 & 1.86 & 22.0 \\
    \method (masked), w/ FT  & \textbf{1.228} & \textbf{.834} & \textbf{.523} & \textbf{1.167} & \textbf{.840} & \textbf{.544} & \textbf{0.76} & \textbf{17.1} \\
    \bottomrule
  \end{tabular}
\end{table*}

\subsection{Distillation}
\label{sec:supp_distill}
\Cref{tab:distill_seg,tab:distill_gt} compare both teachers with their one-step students (\cref{sec:supp_distill_train}).
The students match their teachers in accuracy: CD changes by at most 0.015\,cm, NC improves everywhere, and F@1 is matched or improved.
The students produce somewhat more holes (\eg 1.61$\times$ \vs 1.04$\times$ for \method on 4D-DRESS and 1.49$\times$ \vs 1.00$\times$ on CloSe); such holes can mostly be rectified with standard mesh post-processing such as hole filling~\cite{xiang2025trellis2}.
Reflow keeps the surface cleaner: relative to the student with reflow (1.00$\times$), distilling the teacher directly produces 1.06--1.19$\times$ as many holes on 4D-DRESS and CloSe, at similar accuracy.
Run times are given in \cref{sec:runtime}.

\begin{table}[!htbp]
  \centering
  \caption{Teacher vs.\ distilled student, segmentation protocol.
  4D-DRESS: 64 frames, ours on the GVHMR~\cite{shen2024gvhmr} body; CloSe: 146 frontal frames, ours on the PromptHMR~\cite{wang2025prompthmr} body.
  Holes are relative to the lowest value among the four models on each dataset (\cref{sec:supp_mesh_metrics}).}
  \label{tab:distill_seg}
  \setlength{\tabcolsep}{3.5pt}
  \resizebox{\linewidth}{!}{%
  \begin{tabular}{@{}lccccc@{}}
    \toprule
    Model & CD\,$\downarrow$ & NC\,$\uparrow$ & F@1\,$\uparrow$ & Pen.\,(\%)\,$\downarrow$ & Holes\,$\downarrow$ \\
    \midrule
    \multicolumn{6}{@{}l}{\textit{4D-DRESS}} \\
    \method (teacher)            & 1.484 & .813 & .481 & 1.65 & 1.04$\times$ \\
    \quad Student                 & 1.475 & .823 & .487 & 1.42 & 1.61$\times$ \\
    \method, masked (teacher)    & 1.528 & .811 & .465 & 0.58 & 1.00$\times$ \\
    \quad Student                 & 1.529 & .819 & .466 & 0.46 & 1.56$\times$ \\
    \midrule
    \multicolumn{6}{@{}l}{\textit{CloSe}} \\
    \method (teacher)            & 1.202 & .837 & .534 & 1.25 & 1.00$\times$ \\
    \quad Student                 & 1.196 & .845 & .536 & 1.26 & 1.49$\times$ \\
    \method, masked (teacher)    & 1.228 & .834 & .523 & 0.76 & 1.07$\times$ \\
    \quad Student                 & 1.222 & .841 & .525 & 0.74 & 1.64$\times$ \\
    \bottomrule
  \end{tabular}}
\end{table}

\begin{table}[!htbp]
  \centering
  \caption{Teacher vs.\ distilled student, GT label-transfer protocol.
  Setup as in \cref{tab:distill_seg}.}
  \label{tab:distill_gt}
  \setlength{\tabcolsep}{3.5pt}
  \resizebox{\linewidth}{!}{%
  \begin{tabular}{@{}lccccc@{}}
    \toprule
    Model & CD\,$\downarrow$ & NC\,$\uparrow$ & F@1\,$\uparrow$ & Pen.\,(\%)\,$\downarrow$ & Holes\,$\downarrow$ \\
    \midrule
    \multicolumn{6}{@{}l}{\textit{4D-DRESS}} \\
    \method (teacher)            & 1.477 & .816 & .485 & 1.65 & 1.04$\times$ \\
    \quad Student                 & 1.475 & .823 & .491 & 1.42 & 1.61$\times$ \\
    \method, masked (teacher)    & 1.503 & .815 & .476 & 0.58 & 1.00$\times$ \\
    \quad Student                 & 1.518 & .820 & .476 & 0.46 & 1.56$\times$ \\
    \midrule
    \multicolumn{6}{@{}l}{\textit{CloSe}} \\
    \method (teacher)            & 1.148 & .843 & .553 & 1.25 & 1.00$\times$ \\
    \quad Student                 & 1.144 & .848 & .555 & 1.26 & 1.49$\times$ \\
    \method, masked (teacher)    & 1.167 & .840 & .544 & 0.76 & 1.07$\times$ \\
    \quad Student                 & 1.158 & .846 & .547 & 0.74 & 1.64$\times$ \\
    \bottomrule
  \end{tabular}}
\end{table}

\section{Robustness to Truncation}
\label{sec:supp_occlusion}

\begin{figure*}[!tp]
  \centering
  \includegraphics[width=0.8\linewidth]{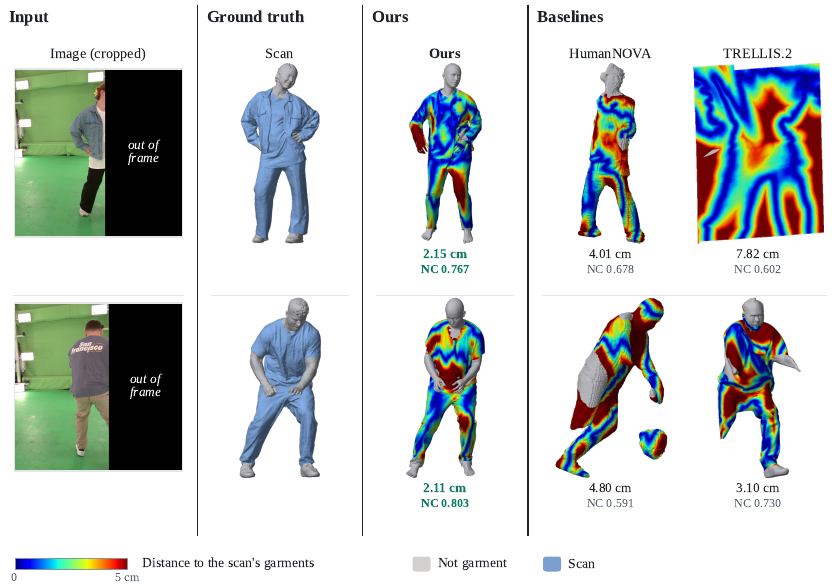}
  \caption{\textbf{Truncated inputs on 4D-DRESS~\cite{wang20244ddress}.}
  Part of the person is out of frame (black).
  Our method still generates a complete garment on the given body, whereas the single-surface methods HumanNOVA~\cite{hu2026humannova} and TRELLIS.2~\cite{xiang2025trellis2}, which must also infer the body, produce fragmented or collapsed surfaces.
  CD and NC are given below each result.}
  \label{fig:occlusion}
\end{figure*}

\Cref{fig:occlusion} evaluates inputs in which a large part of the person lies outside the image.
Because the body is an input to our method, its pose and extent are known even where the image is missing, and the model completes the garment on the unseen side.
The single-surface baselines have no such anchor: HumanNOVA hallucinates detached parts, and TRELLIS.2 can collapse to a flat plane.

\section{Per-Garment Generation}
\label{sec:supp_pergarment}

\begin{figure*}[!tp]
  \centering
  \includegraphics[width=\linewidth]{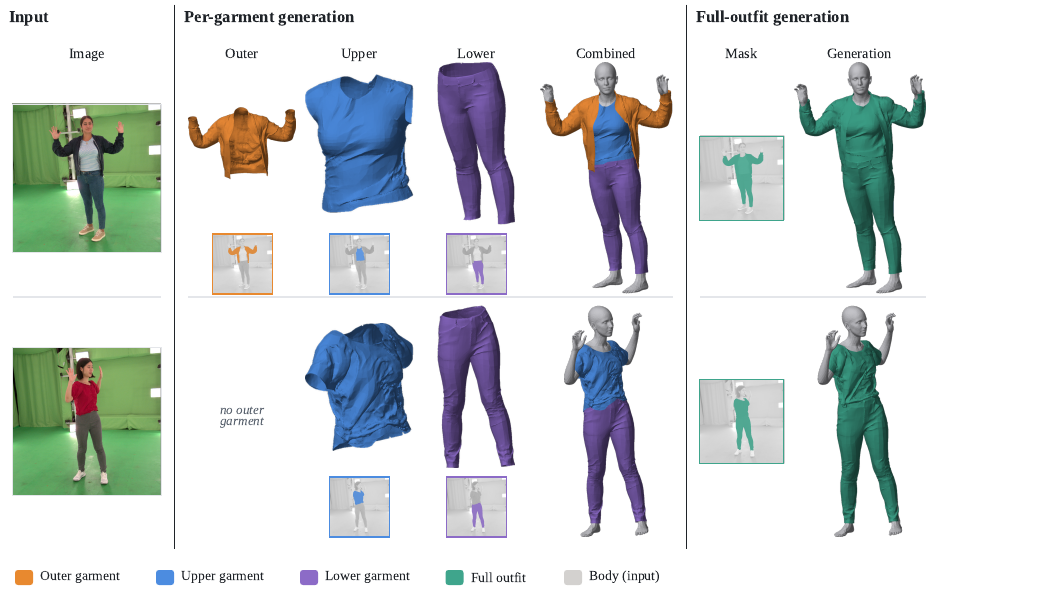}
  \caption{\textbf{Per-garment generation from garment masks} on 4D-DRESS~\cite{wang20244ddress}.
  A mask selects the garment and the same model generates it; garments generated separately (outer, upper, lower) combine on one body after layer resolve.
  Given the full-outfit mask, the model instead generates the whole outfit at once.
  Masks are taken from the scan's garment labels in the input crop, with one run per mask.}
  \label{fig:pergarment}
\end{figure*}

\Cref{fig:pergarment} shows per-garment generation from garment masks: the same model generates individual garments or the complete outfit, and garments generated in separate runs assemble into a layered outfit after layer resolve (\cref{sec:supp_layer}).

\subsection{Robustness to the Garment Mask}
\label{sec:supp_mask_robust}
Since the masked model generates the garment its mask selects, we test how errors in the mask propagate.
On the 64 4D-DRESS frames, we keep the model and the GVHMR body fixed and change only the mask: synthetic errors applied to the ground-truth mask (dilation, erosion, a jagged boundary, and random holes removing 25\% of the mask area), and automatic masks from Grounded SAM~\cite{ren2024groundedsam} with one text prompt per garment class.
Each condition is compared with the ground-truth mask on the same garments with a paired Wilcoxon signed-rank test (Holm-corrected).
No difference is significant (\cref{tab:mask_robust}): even a mask enlarged to an IoU of 0.51 changes the garment CD by at most 0.006\,cm.

\begin{table}[!tbp]
\centering\small
\setlength{\tabcolsep}{4pt}
\begin{tabular}{@{}lcccc@{}}
\toprule
Conditioning mask & IoU & $\Delta$CD (cm) & $\Delta$NC & $\Delta$F@1 \\
\midrule
Dilate 4\,px       & 0.81 & $-0.001$ & 0.000 & $+0.001$ \\
Dilate 16\,px      & 0.51 & $+0.005$ & 0.000 & $-0.001$ \\
Erode 4\,px        & 0.78 & $+0.001$ & 0.000 & $+0.002$ \\
Jagged boundary    & 0.72 & $+0.004$ & 0.000 & 0.000 \\
25\% holes         & 0.73 & $+0.006$ & 0.000 & $-0.002$ \\
\midrule
Grounded SAM       & 0.94 & $-0.005$ & 0.000 & 0.000 \\
\bottomrule
\end{tabular}
\caption{\textbf{Robustness to the garment mask} (masked model, 4D-DRESS, per garment).
Change with respect to the ground-truth mask on the same garments (lower CD and higher NC\,/\,F@1 are better); no difference is significant ($p\geq0.05$, paired Wilcoxon, Holm-corrected).
Reference with the ground-truth mask: 1.45\,cm / 0.824 / 0.484 (\cref{tab:layers_4dd}).
IoU: mean IoU to the ground-truth mask.}
\label{tab:mask_robust}
\end{table}

\section{Layered Garment Evaluation}
\label{sec:supp_layered}

Each garment of a multi-garment outfit is generated separately, with one run per garment mask, and scored on its own against the corresponding ground-truth garment.
\Cref{tab:layers_bedlam} compares with the baselines on BEDLAM2.0~\cite{tesch2025bedlam2}, where the simulated garments are complete, including the parts hidden by other garments.
\Cref{tab:layers_4dd} reports our per-garment accuracy on 4D-DRESS~\cite{wang20244ddress}, where the scans hold only the visible surface.

\begin{table}[!tbp]
\centering
\small
\setlength{\tabcolsep}{3.5pt}
\resizebox{\linewidth}{!}{%
\begin{tabular}{lccc|ccc}
\toprule
 & \multicolumn{3}{c|}{Complete layer} & \multicolumn{3}{c}{Visible part} \\
Method & CD$\downarrow$ & NC$\uparrow$ & F@1cm$\uparrow$ & CD$\downarrow$ & NC$\uparrow$ & F@1cm$\uparrow$ \\
\midrule
ReMu & 2.45 & 0.785 & 0.340 & 2.44 & 0.781 & 0.331 \\
ISP  & 2.54 & 0.809 & 0.327 & 2.52 & 0.806 & 0.317 \\
Ours & \textbf{1.26} & \textbf{0.852} & \textbf{0.566} & \textbf{1.27} & \textbf{0.849} & \textbf{0.555} \\
\bottomrule
\end{tabular}}
\caption{\textbf{Per-garment evaluation on the BEDLAM2.0 validation split} (200 frames from held-out validation sequences). Each garment of a multi-garment outfit is scored on its own. Ours: one generation per garment mask. Baselines (ReMu~\cite{vuran2025remu}, ISP~\cite{li2023isp}): each GT garment is matched to the baseline's best layer
(oracle assignment). \emph{Complete}: against the whole simulated garment, including what other garments hide in the photo.
CD in cm.}
\label{tab:layers_bedlam}
\end{table}

\begin{table}[!tbp]
\centering
\small
\setlength{\tabcolsep}{4pt}
\resizebox{\linewidth}{!}{%
\begin{tabular}{lcccccc}
\toprule
Garment & n & CD$\downarrow$ & NC$\uparrow$ & F@1cm$\uparrow$ & Hidden & Pen.$\downarrow$ \\
\midrule
Upper, no outer layer         & 31 & 1.20 & 0.835 & 0.539 &  5\% & 0.42 \\
Upper, under an outer layer   & 20 & 1.44 & 0.811 & 0.470 & 71\% & 2.76 \\
Lower, Inner takes            & 30 & 1.14 & 0.875 & 0.560 & 13\% & 1.69 \\
Lower, Outer takes            & 29 & 1.41 & 0.859 & 0.468 & 20\% & 0.47 \\
Outer                         & 32 & 2.02 & 0.743 & 0.382 &  8\% & 0.60 \\
\midrule
All layers                    & 142 & 1.45 & 0.824 & 0.484 & 20\% & 1.07 \\
\textit{Whole outfit, same frames} & \textit{64} & \textit{1.50} & \textit{0.815} & \textit{0.476} & -- & \textit{0.58} \\
\bottomrule
\end{tabular}}
\caption{\textbf{Per-garment evaluation on 4D-DRESS.} One generation per garment label (upper / lower / outer), scored against
that label of the scan. Scans hold only the visible surface, so the part generated under another garment is not scored; \emph{Hidden}
is its share of the generated garment's surface. CD in cm; Pen.: \% of garment area inside the body.}
\label{tab:layers_4dd}
\end{table}

\section{Garment Retargeting}
\label{sec:supp_transfer}

\begin{figure*}[!tp]
  \centering
  \includegraphics[width=\linewidth]{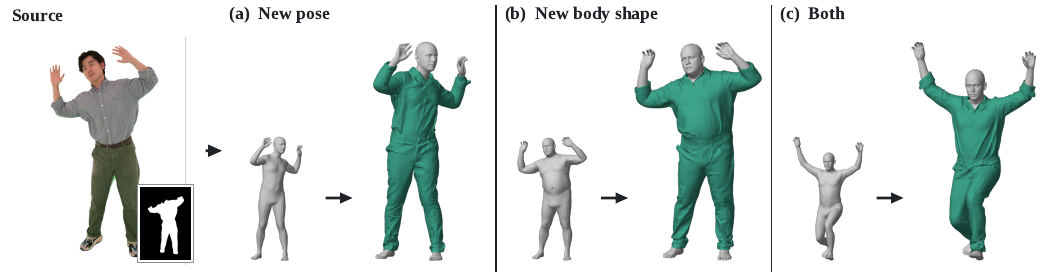}
  \caption{\textbf{Garment retargeting} across poses and body shapes. Keeping the source image and garment mask fixed, our method generates the garments on target bodies with (a) a new pose, (b) a new body shape, and (c) both. All results use the same trained model.}
  \label{fig:transfer}
\end{figure*}

\Cref{fig:transfer} shows garment retargeting to a new pose, a new body shape, and both at once.
\Cref{tab:xfer} evaluates it quantitatively on BEDLAM2.0: the outfit in a source photo is generated directly on another person's body and compared with the same outfit simulated in that person's size and pose.

\begin{table}[!tbp]
\centering
\small
\setlength{\tabcolsep}{4pt}
\resizebox{\linewidth}{!}{%
\begin{tabular}{lcccc}
\toprule
Method & CD$\downarrow$ & NC$\uparrow$ & F@1cm$\uparrow$ & Pen.$\downarrow$ \\
\midrule
Ours, generated directly on the target body & \textbf{0.75} & \textbf{0.886} & \textbf{0.808} & \textbf{1.09} \\
Ours on the source + retarget               & 1.10 & 0.840 & 0.706 & 5.76 \\
Source GT + retarget (oracle)               & 1.06 & 0.823 & 0.708 & 5.94 \\
\midrule
\textit{Ours from the target's own photo (upper bound)} & \textit{0.56} & \textit{0.905} & \textit{0.899} & \textit{0.98} \\
\bottomrule
\end{tabular}}
\vspace{4pt}

\resizebox{\linewidth}{!}{%
\begin{tabular}{lcccccc}
\toprule
 & \multicolumn{3}{c}{$|\Delta$size$|$ (steps)} & \multicolumn{2}{c}{Pose gap} & \\
\cmidrule(lr){2-4}\cmidrule(lr){5-6}
CD$\downarrow$ & 1 & 2 & $\geq$3 & low & high & All \\
\midrule
Ours, direct            & \textbf{0.74} & \textbf{0.74} & \textbf{0.77} & \textbf{0.75} & \textbf{0.76} & \textbf{0.75} \\
Ours + retarget         & 1.07 & 1.07 & 1.14 & 1.02 & 1.17 & 1.10 \\
Source GT + retarget    & 1.02 & 1.03 & 1.11 & 0.97 & 1.14 & 1.06 \\
\midrule
Pairs                   & 420 & 317 & 463 & 400 & 400 & 1200 \\
\bottomrule
\end{tabular}}
\caption{\textbf{Generating garments on a different body.} BEDLAM2 validation: the outfit in a source photo is generated on another
person's body, and compared with the same outfit simulated in that person's size and pose. Direct generation beats reconstructing on
the source and retargeting on 94\% of pairs, even when the source garment is perfect (oracle). Pose gap: PA-MPJPE terciles
($\leq$103\,mm / $>$145\,mm). CD in cm, F-score at 1\,cm, Pen.: \% of garment area inside the body.}
\label{tab:xfer}
\end{table}

\section{Garment Animation}
\label{sec:supp_anim}

This section details the animation experiment of \cref{tab:animation} and gives its full results, with our A-pose garment as a second two-stage source (\cref{tab:animation_full}): the protocol, how each baseline is run, the run time of DrapeNet and HOOD, and the results without smoothing the binding weights.

\subsection{Protocol}
\label{sec:supp_anim_protocol}

\subsubsection{Task}
The task is to produce the subject's garments in a new target pose, scored against the garment actually worn in the target frame.
In the \emph{two-stage} route, a garment in a canonical A-pose is bound to the body and moved to the target pose by a retargeting method (LBS~\cite{magnenat1988lbs}, DrapeNet~\cite{deluigi2023drapenet} or HOOD~\cite{grigorev2023hood}).
The A-pose garment comes from one of two sources.
\emph{Ours}: our model, conditioned on the input image, a garment mask and the subject's body in the A-pose.
\emph{Ground truth (GT)}: the real garment of the same subject and outfit; on 4D-DRESS, the scan garment from an A-pose-like frame of the same take, moved onto the A-pose body; on BEDLAM2.0, the graded A-pose garment of the simulated outfit.
In the \emph{direct} route, our model generates the garment on the target body in one pass, with no binding, skinning or simulation.

\subsubsection{Benchmarks}
On 4D-DRESS~\cite{wang20244ddress} we use the 64 frames of \cref{tab:main} (32 Inner and 32 Outer outfits); target bodies are GVHMR~\cite{shen2024gvhmr} estimates, our input image is taken from the same take at least 20 frames away from the target frame, and garments are scored against the label-transferred target scan garment with the alignment of \cref{tab:main}.
On BEDLAM2.0~\cite{tesch2025bedlam2} we use the 200 validation frames; target bodies are estimated from the target frame with PromptHMR~\cite{wang2025prompthmr}, our input image is a different frame of the same sequence, and the simulated target garment is the ground truth.
Twelve BEDLAM2.0 frames are excluded because the subject crouches or bends and a standing A-pose body does not fit in the image crop at the crop's scale, which leaves 188.
All metrics are as in \cref{sec:supp_eval}: CD (cm), NC and F@1.

\subsubsection{Excluded Frames}
Frames are always left out for both garment sources, so each row is paired.
HOOD on 4D-DRESS runs on the 32 Inner-outfit frames, since it does not model the cloth--cloth contact of open outerwear (\cref{sec:supp_animation}).
A HOOD frame fails when more than 5\% of the garment ends up over 10\,cm from the body, and failures of either source are excluded from that row.
DrapeNet on BEDLAM2.0 covers the 101 outfits with exactly one top and one bottom, the only case it supports.
In \cref{tab:animation,tab:animation_full}, each row is thus scored on 64 frames for LBS and DrapeNet and 28 / 27 for HOOD static / motion on 4D-DRESS (23 / 24 with the nearest-vertex binding of \cref{tab:animation_nn}), and on 188 for LBS, 101 for DrapeNet and 157 / 151 for HOOD static / motion on BEDLAM2.0; the direct column is scored on exactly the frames of its row.
Before scoring HOOD, detached pieces whose median distance to the body exceeds 10\,cm are removed.

\begin{table*}[!tp]
  \centering
  \caption{\textbf{Garment animation on 4D-DRESS~\cite{wang20244ddress} and BEDLAM2.0~\cite{tesch2025bedlam2}, full results} (\cref{tab:animation} with our A-pose garment as a second two-stage source; validation split; CD in cm against the garment worn in the target frame).
  Two-stage routes bind an A-pose garment, either the ground-truth (GT) garment of the same outfit or our A-pose generation, and move it to the target pose with LBS, DrapeNet~\cite{deluigi2023drapenet} or HOOD~\cite{grigorev2023hood}; \method direct instead generates the garment on the target body in one pass.
  Results with unsmoothed binding are in \cref{tab:animation_nn}.}
  \label{tab:animation_full}
  \small
  \setlength{\tabcolsep}{4pt}
  \begin{tabular}{@{}l ccc ccc ccc@{}}
    \toprule
    & \multicolumn{3}{c}{GT garment + retargeting} & \multicolumn{3}{c}{\method A-pose + retargeting} & \multicolumn{3}{c}{\method direct} \\
    \cmidrule(lr){2-4}\cmidrule(lr){5-7}\cmidrule(l){8-10}
    Retargeting & CD\,$\downarrow$ & NC\,$\uparrow$ & F@1\,$\uparrow$ & CD\,$\downarrow$ & NC\,$\uparrow$ & F@1\,$\uparrow$ & CD\,$\downarrow$ & NC\,$\uparrow$ & F@1\,$\uparrow$ \\
    \midrule
    \multicolumn{10}{@{}l}{\textit{4D-DRESS}} \\
    \quad LBS (per garment)                     & \underline{1.742} & \textbf{.815} & \underline{.430} & 1.839 & .801 & .426 & \textbf{1.681} & \underline{.806} & \textbf{.446} \\
    \quad LBS (whole outfit)                    & --    & --   & --   & \underline{1.782} & \underline{.804} & \underline{.426} & \textbf{1.681} & \textbf{.806} & \textbf{.446} \\
    \quad DrapeNet~\cite{deluigi2023drapenet}   & \underline{1.823} & \underline{.804} & .410 & 1.838 & .799 & \underline{.414} & \textbf{1.681} & \textbf{.806} & \textbf{.446} \\
    \quad HOOD~\cite{grigorev2023hood}, static  & 1.492 & \underline{.849} & .440 & \underline{1.475} & .844 & \underline{.456} & \textbf{1.218} & \textbf{.857} & \textbf{.532} \\
    \quad HOOD~\cite{grigorev2023hood}, motion  & 1.503 & \underline{.846} & .435 & \underline{1.475} & .842 & \underline{.453} & \textbf{1.202} & \textbf{.860} & \textbf{.536} \\
    \midrule
    \multicolumn{10}{@{}l}{\textit{BEDLAM2.0}} \\
    \quad LBS (per garment)                     & \underline{0.992} & .862 & \underline{.739} & 1.071 & \underline{.864} & .689 & \textbf{0.631} & \textbf{.891} & \textbf{.889} \\
    \quad LBS (whole outfit)                    & --    & --   & --   & \underline{1.088} & \underline{.861} & \underline{.685} & \textbf{0.631} & \textbf{.891} & \textbf{.889} \\
    \quad DrapeNet~\cite{deluigi2023drapenet}   & \underline{1.152} & \underline{.847} & \underline{.633} & 1.232 & .839 & .605 & \textbf{0.616} & \textbf{.886} & \textbf{.901} \\
    \quad HOOD~\cite{grigorev2023hood}, static  & \underline{1.088} & \underline{.852} & \underline{.694} & 1.200 & .836 & .667 & \textbf{0.621} & \textbf{.893} & \textbf{.893} \\
    \quad HOOD~\cite{grigorev2023hood}, motion  & \underline{1.169} & \underline{.845} & \underline{.666} & 1.289 & .831 & .643 & \textbf{0.617} & \textbf{.894} & \textbf{.895} \\
    \bottomrule
  \end{tabular}
\end{table*}

\subsection{Methods}
\label{sec:supp_anim_methods}

To ensure a fair comparison, we made every reasonable effort to run each baseline under the most favourable conditions available to it.
Third-party methods run with their released code and weights, unchanged, and all glue code (body fitting, binding, moving garments between bodies) is shared by both garment sources.
Where a component was missing or failed, we supplied the most faithful substitute we could, such as fitting DrapeNet's latent codes through its released decoder, and we smooth the binding (\cref{sec:supp_anim_smooth}), which benefits the ground-truth garments most.

\subsubsection{Our Garments}
A-pose garments are generated by the masked fine-tuned model with default post-processing (\cref{sec:supp_postproc}), conditioned on the image, the mask and four renders of the subject's A-pose body.
We run one generation per garment mask, inner to outer, plus one with the whole-outfit mask.
Direct generation uses the same model with the target body as the conditioning body: one forward pass per garment, with no binding or simulation.

\subsubsection{Binding}
Shared by all retargeting methods, each garment vertex takes the SMPL-X skinning weights of its four nearest body vertices, weighted by inverse distance.
The same binding unposes garments into each retargeting method's canonical pose and moves garments between bodies (shape offsets from the nearest body vertex).
\Cref{tab:animation} smooths this binding over the garment mesh (\cref{sec:supp_anim_smooth}); \cref{tab:animation_nn} uses it as is.

\subsubsection{\texorpdfstring{LBS~\cite{magnenat1988lbs}}{LBS}}
The bound garment follows the body's joint transforms from the A-pose to the target pose.
We report it per garment (the same top and bottom DrapeNet uses) and for the whole outfit at once; the latter only for our garments, since the GT source is split per garment.

\subsubsection{\texorpdfstring{DrapeNet~\cite{deluigi2023drapenet}}{DrapeNet}}
DrapeNet drapes on a female SMPL body.
We fit one to each SMPL-X body (joints and surface) in an upright world frame and carry the result back onto the SMPL-X surface.
The garment is unposed with the binding, moved to DrapeNet's neutral shape by the nearest-vertex shape offset, and posed to DrapeNet's canonical pose.
Since the garment encoder is not released, each garment's 32-d latent code is fitted through the released UDF decoder (surface and near-surface points, with a small pull towards the mean training code).

\subsubsection{\texorpdfstring{HOOD~\cite{grigorev2023hood}}{HOOD}}
Following the official inference notebook and its material settings, on the same female SMPL fit as for DrapeNet, carried back to SMPL-X.
Each garment is unposed to HOOD's rest pose (zero pose, zero shape) with the binding and remeshed to about 10k vertices, HOOD's own working resolution (its garments have 4--12k vertices).
The top 3\,cm of each bottom garment (the waistband) is pinned; the official notebook selects these vertices by hand.
The sequence starts with HOOD's own 30-frame interpolation from its rest pose to the first pose.
\emph{Static}: the body is then interpolated from the A-pose to the target pose over 40 frames and held for 20 frames to let the cloth settle.
\emph{Motion}: the body follows the real motion leading to the target frame; on 4D-DRESS, GVHMR estimates over the take from its first frame, on BEDLAM2.0 the sequence's own motion.
Simulation runs in the dataset's upright world frame, so gravity points down, and the last frame is scored.

\subsection{Run Time and Paired Comparison}
\label{sec:supp_anim_runtime}

\begin{table}[!htbp]
  \centering
  \caption{\textbf{Run time per frame of DrapeNet and HOOD} (median over the 4D-DRESS frames, one NVIDIA H200 shared between jobs).
  Retargeting methods run on top of the A-pose generation, which is not included in their rows.}
  \label{tab:anim_runtime}
  \setlength{\tabcolsep}{3pt}
  \resizebox{\linewidth}{!}{%
  \begin{tabular}{@{}lll@{}}
    \toprule
    Route & Per frame & Dominated by \\
    \midrule
    DrapeNet & 95--110\,s & SMPL fit, latent-code fit \\
    DrapeNet, smoothed binding & $\approx$120\,s & as above, plus smoothing \\
    HOOD, static & 3.5--4.5\,min & simulating $\approx$90 frames \\
    HOOD, motion & 6--7\,min & simulating the take's motion \\
    \bottomrule
  \end{tabular}}
\end{table}

\Cref{tab:anim_runtime} gives the cost of the learned retargeting methods, which run on top of generating the A-pose garments.

\begin{table}[!htbp]
  \centering
  \caption{\textbf{Paired CD gain of \method direct over the GT-garment route on 4D-DRESS} (cm, positive: direct is better), with 95\% confidence intervals and the number of frames on which direct generation has the lower CD.}
  \label{tab:anim_paired}
  \setlength{\tabcolsep}{3pt}
  \resizebox{\linewidth}{!}{%
  \begin{tabular}{@{}lccc ccc@{}}
    \toprule
    & \multicolumn{3}{c}{Nearest-vertex binding} & \multicolumn{3}{c}{Smoothed binding} \\
    \cmidrule(lr){2-4}\cmidrule(l){5-7}
    Against & Gain & 95\% CI & Won & Gain & 95\% CI & Won \\
    \midrule
    GT + LBS          & 0.24 & 0.11 to 0.36 & 46/64 & 0.06 & $-$0.02 to 0.14 & 42/64 \\
    GT + DrapeNet     & 0.26 & 0.14 to 0.38 & 52/64 & 0.14 & 0.06 to 0.23 & 46/64 \\
    GT + HOOD, static & 0.37 & 0.30 to 0.46 & 23/23 & 0.27 & 0.20 to 0.35 & 27/28 \\
    GT + HOOD, motion & 0.45 & 0.35 to 0.56 & 24/24 & 0.30 & 0.23 to 0.38 & 27/27 \\
    \bottomrule
  \end{tabular}}
\end{table}

\Cref{tab:anim_paired} gives the paired gains behind \cref{tab:animation_nn} (nearest-vertex binding) and \cref{tab:animation} (smoothed binding).
With nearest-vertex binding, direct generation is significantly better than every GT route and wins on every HOOD frame.

\subsection{Smoothed vs.\ Nearest-Vertex Binding}
\label{sec:supp_anim_smooth}

Plain nearest-vertex binding assigns unrelated weights to neighbouring garment vertices that lie close to different body parts, \eg at the armpit, so the garment tears as the body moves; closely fitting scan garments are affected most.
We therefore diffuse every bound field $w$ (skinning weights and shape offsets) over the garment mesh for 50 rounds,
\begin{equation}
  w_i \leftarrow \tfrac{1}{2}\, w_i + \tfrac{1}{2}\, \frac{1}{|\mathcal{N}(i)|} \sum_{j \in \mathcal{N}(i)} w_j,
\end{equation}
where $\mathcal{N}(i)$ are the mesh neighbours of vertex $i$, for both garment sources and all retargeting methods; coincident seam vertices are merged beforehand so that seams remain closed.

\begin{table}[!htbp]
  \centering
  \caption{\textbf{Torn area before $\rightarrow$ after smoothing the binding}: share of the garment area in faces with an edge over 5\,cm. HOOD on BEDLAM2.0 was run only with smoothed binding.}
  \label{tab:anim_torn}
  \setlength{\tabcolsep}{3pt}
  \resizebox{\linewidth}{!}{%
  \begin{tabular}{@{}lcccc@{}}
    \toprule
    & \multicolumn{2}{c}{4D-DRESS} & \multicolumn{2}{c}{BEDLAM2.0} \\
    \cmidrule(lr){2-3}\cmidrule(l){4-5}
    Retargeting & GT & Ours & GT & Ours \\
    \midrule
    LBS (per garment) & 18.8\% $\rightarrow$ 0.9\% & 4.4\% $\rightarrow$ 0.8\% & 3.4\% $\rightarrow$ 1.2\% & 2.4\% $\rightarrow$ 0.0\% \\
    DrapeNet          & 21.8\% $\rightarrow$ 1.3\% & 6.5\% $\rightarrow$ 0.2\% & 4.0\% $\rightarrow$ 0.1\% & 5.1\% $\rightarrow$ 0.1\% \\
    HOOD, static      & 12.8\% $\rightarrow$ 0.2\% & 4.6\% $\rightarrow$ 1.4\% & -- $\rightarrow$ 1.4\% & -- $\rightarrow$ 2.4\% \\
    \bottomrule
  \end{tabular}}
\end{table}

\begin{table*}[!tp]
  \centering
  \caption{\textbf{Garment animation with nearest-vertex binding weights}, without smoothing.
  Same protocol as \cref{tab:animation}, with the skinning weights and shape offsets of the four nearest body vertices used as is, for both garment sources.
  HOOD on BEDLAM2.0 was run only with smoothed binding.}
  \label{tab:animation_nn}
  \small
  \setlength{\tabcolsep}{4pt}
  \begin{tabular}{@{}l ccc ccc ccc@{}}
    \toprule
    & \multicolumn{3}{c}{GT garment + retargeting} & \multicolumn{3}{c}{\method A-pose + retargeting} & \multicolumn{3}{c}{\method direct} \\
    \cmidrule(lr){2-4}\cmidrule(lr){5-7}\cmidrule(l){8-10}
    Retargeting & CD\,$\downarrow$ & NC\,$\uparrow$ & F@1\,$\uparrow$ & CD\,$\downarrow$ & NC\,$\uparrow$ & F@1\,$\uparrow$ & CD\,$\downarrow$ & NC\,$\uparrow$ & F@1\,$\uparrow$ \\
    \midrule
    \multicolumn{10}{@{}l}{\textit{4D-DRESS}} \\
    \quad LBS (per garment)                     & 1.917 & .788 & .407 & \underline{1.886} & \underline{.791} & \underline{.421} & \textbf{1.681} & \textbf{.806} & \textbf{.446} \\
    \quad LBS (whole outfit)                    & --    & --   & --   & \underline{1.838} & \underline{.793} & \underline{.419} & \textbf{1.681} & \textbf{.806} & \textbf{.446} \\
    \quad DrapeNet~\cite{deluigi2023drapenet}   & 1.941 & .779 & .395 & \underline{1.922} & \underline{.782} & \underline{.403} & \textbf{1.681} & \textbf{.806} & \textbf{.446} \\
    \quad HOOD~\cite{grigorev2023hood}, static  & \underline{1.571} & .830 & .430 & 1.586 & \underline{.834} & \underline{.444} & \textbf{1.201} & \textbf{.861} & \textbf{.543} \\
    \quad HOOD~\cite{grigorev2023hood}, motion  & 1.645 & .826 & .419 & \underline{1.563} & \underline{.831} & \underline{.435} & \textbf{1.197} & \textbf{.860} & \textbf{.540} \\
    \midrule
    \multicolumn{10}{@{}l}{\textit{BEDLAM2.0}} \\
    \quad LBS (per garment)                     & \underline{1.009} & \underline{.853} & \underline{.734} & 1.092 & .852 & .683 & \textbf{0.631} & \textbf{.891} & \textbf{.889} \\
    \quad LBS (whole outfit)                    & --    & --   & --   & \underline{1.110} & \underline{.848} & \underline{.680} & \textbf{0.631} & \textbf{.891} & \textbf{.889} \\
    \quad DrapeNet~\cite{deluigi2023drapenet}   & \underline{1.191} & \underline{.832} & \underline{.618} & 1.328 & .813 & .578 & \textbf{0.616} & \textbf{.886} & \textbf{.901} \\
    \bottomrule
  \end{tabular}
\end{table*}

Smoothing reduces the torn area to at most 1.4\% (\cref{tab:anim_torn}) and benefits the ground-truth garments most: with it, the GT garment becomes the strongest two-stage source, except with HOOD on 4D-DRESS, where both sources are on par; without it (\cref{tab:animation_nn}), our A-pose garment is on par with or better than the GT garment for most retargeting methods on 4D-DRESS.
Direct generation remains ahead with either binding; its lead over GT + LBS on 4D-DRESS decreases from 0.24\,cm to 0.06\,cm with smoothing (\cref{tab:anim_paired}).

\section{Statistical Significance}
\label{sec:supp_significance}

We test all comparisons with a paired, two-sided Wilcoxon signed-rank test, Holm-corrected for multiple comparisons. Our improvements are significant against every baseline on 4D-DRESS, in the wild, per garment and in cross-body retargeting. On CloSe, the differences to HumanNOVA, and TRELLIS.2 in CD and F@1, are not significant. In animation, direct generation is significantly better than every two-stage route. (\cref{tab:anim_paired}).
\section{Qualitative Animation Results}
\label{sec:supp_animation}

\begin{figure*}[!tp]
  \centering
  \includegraphics[width=\linewidth]{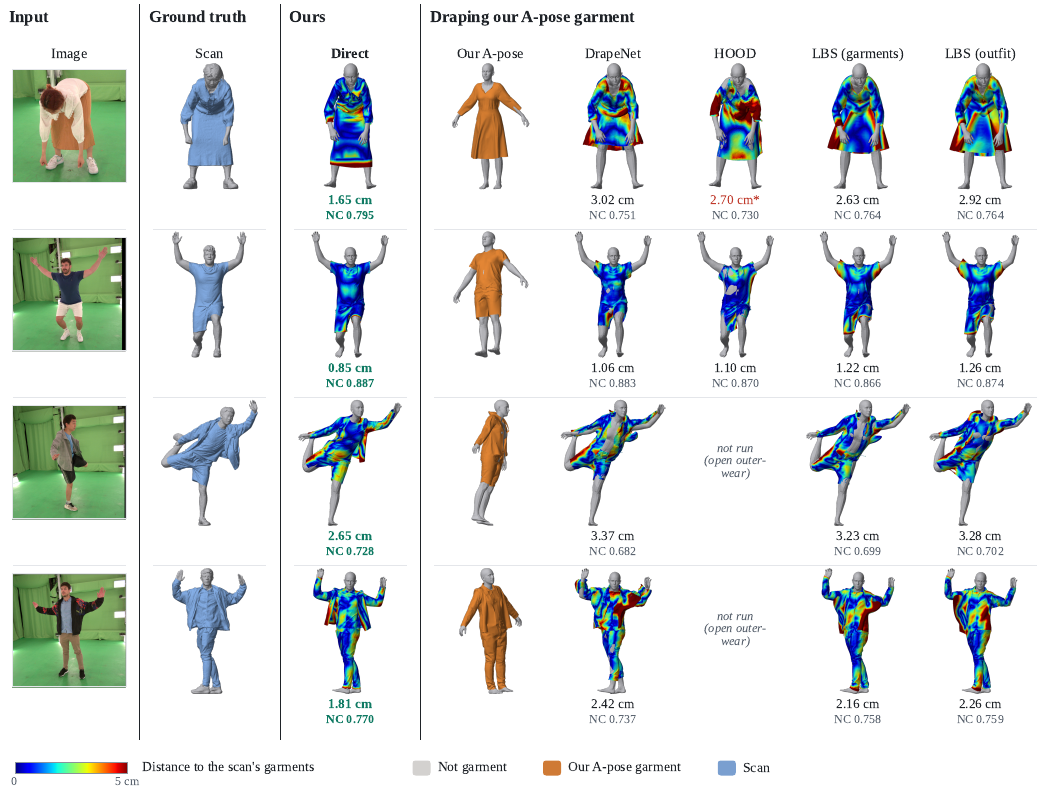}
  \caption{\textbf{Animation on 4D-DRESS~\cite{wang20244ddress}.}
  The garment is read from the source image and generated directly on the body of the target frame (Direct); the baselines repose the A-pose garment with DrapeNet~\cite{deluigi2023drapenet}, HOOD~\cite{grigorev2023hood}, or LBS applied per garment or to the whole outfit.
  Garments are colored by distance to the target scan (0--5\,cm); CD and NC are given below each result.
  HOOD is run on inner outfits only, since it does not handle open outerwear.
  $^{*}$\,HOOD failure on this frame (more than 5\% of the garment ends up over 10\,cm from the body); such frames are excluded from \cref{tab:animation_full} and are shown here only for illustration.}
  \label{fig:animation}
\end{figure*}

\Cref{fig:animation} compares direct generation with the retargeting baselines on 4D-DRESS.

Because the two-stage baselines deform a rest-pose garment into the target pose, they suffer from stretching artifacts (\cref{fig:animation}).
When the subject bends over, DrapeNet~\cite{deluigi2023drapenet} and LBS stretch the hem of a dress outwards instead of letting it hang, and under large arm and leg motions they stretch open jackets away from the body (red regions).
HOOD~\cite{grigorev2023hood}, trained on closed template garments, resolves cloth--body contact but not cloth--cloth contact, so it is not run on open outerwear.
Generating the garment directly on each target body avoids all of these, since no rest-pose mesh is deformed.

\section{Video Reconstruction}
\label{sec:supp_video}

\begin{figure*}[!tp]
  \centering
  \includegraphics[width=\linewidth]{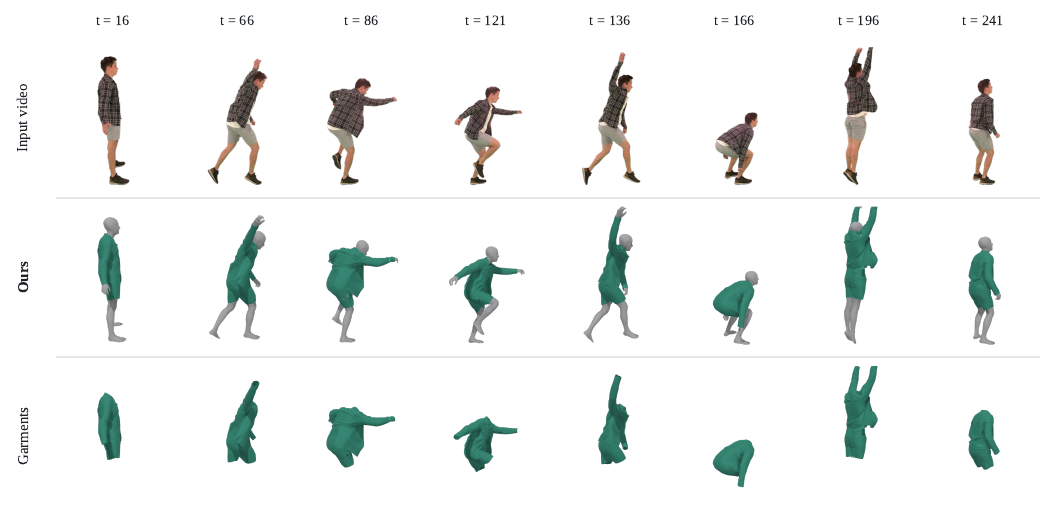}
  \caption{\textbf{Reconstruction from video}, frame by frame.
  Top: input frames at the given time steps $t$.
  Middle: our garments on the estimated body.
  Bottom: the garments alone.
  Each frame is generated independently, without temporal smoothing.}
  \label{fig:video}
\end{figure*}

\Cref{fig:video} applies our method to a video sequence frame by frame.
Each frame is reconstructed independently from its image and the body estimated for that frame, without temporal smoothing.
The sequence contains fast and extreme motion, including running, jumping with raised arms and a deep squat.
The open shirt follows the body through all of these poses, stays outside the body, and keeps its overall shape and length across frames, although it is never constrained to be consistent over time.

\subsection{Temporally Consistent Sampling}
\label{sec:supp_temporal}
For video input, each frame is still generated independently, but to keep consecutive frames from flickering we make two changes to sampling, without retraining.
First, \emph{shared noise}: the sparse-structure stage starts every frame from the same noise draw.
In the SLat stage, the noise is taken from a fixed $64^3$ noise field and indexed by voxel position, so a voxel that is occupied in several frames gets the same noise in each of them.
Second, \emph{latent blending}: all frames are denoised in lockstep, \ie Euler step $k$ is applied to every frame before step $k{+}1$.
After each of the first 9 of the 12 Euler steps, we replace each frame's intermediate latent with a weighted average over a 5-frame window centred on it, using triangular weights $(1,2,3,2,1)/9$.
For the sparse structure this is the dense latent; for the SLat it is done voxel by voxel, averaging only over the frames that occupy that voxel.
The last 3 steps run per frame without blending, so every decoded garment is a plain model output, and the method stays a post-hoc sampling scheme rather than mesh smoothing.

\section{Texturing}
\label{sec:supp_texture}
Our method generates garment geometry; a texture can be added in two simple ways.
\emph{Median color}: each garment gets a single color, the median of the image over its mask.
\emph{TRELLIS.2 texturing}: the garment pixels and our generated mesh are passed to the pretrained texturing stage of TRELLIS.2~\cite{xiang2025trellis2}, which predicts a full texture for the given geometry.